\documentclass{article}

\PassOptionsToPackage{numbers, compress}{natbib}
\usepackage[final]{neurips_2021}
\usepackage{xcolor}
\usepackage{docmute}

\usepackage{neurips_2021}
\usepackage{default}
\usepackage{mystyle}

\title{\pname : Conditional Score-based Diffusion Models for Probabilistic Time Series Imputation}

\author{
  Yusuke Tashiro\textsuperscript{123*}, Jiaming Song\textsuperscript{1}, Yang Song\textsuperscript{1}, Stefano Ermon\textsuperscript{1}\\
  \textsuperscript{1}Department of Computer Science, Stanford University, Stanford, CA, USA\\
  \textsuperscript{2}Mitsubishi UFJ Trust Investment Technology Institute, Tokyo, Japan\\
  \textsuperscript{3}Japan Digital Design, Tokyo, Japan\\
  \texttt{\{ytashiro,tsong,songyang,ermon\}@cs.stanford.edu}
}

\begin{document}

\maketitle

\begin{abstract}
The imputation of missing values in time series has many applications in healthcare and finance. 
While autoregressive models are natural candidates for time series imputation, score-based diffusion models have recently outperformed existing counterparts including autoregressive models in many tasks such as image generation and audio synthesis, and would be promising for time series imputation.
In this paper, we propose \pnamelong{} (\pname), a novel time series imputation method that utilizes score-based diffusion models conditioned on observed data. 
Unlike existing score-based approaches, the conditional diffusion model is explicitly trained for imputation and can exploit correlations between observed values. 
On healthcare and environmental data,  
\pname{} improves by 40-65\% over existing probabilistic imputation methods on popular performance metrics. In addition, deterministic imputation by \pname{} reduces the error by 5-20\% compared to the state-of-the-art deterministic imputation methods. 
Furthermore, \pname{} can also be applied to time series interpolation and probabilistic forecasting, and is competitive with existing baselines. The code is available at \url{https://github.com/ermongroup/CSDI}.
\end{abstract}

\section{Introduction}
\label{sec:introduction}
Multivariate time series are abundant in real world applications such as finance, meteorology and healthcare. These time series data often contain missing values due to various reasons, including device failures and human errors~\citep{silva2012physionet,yi2016st,tan2013traffic}. 
Since missing values 
can hamper the interpretation of a time series, many studies have addressed the task of imputing missing values 
using machine learning techniques~\citep{nelwamondo2007missing,hudak2008nearest,buuren2010mice}. 
In the past few years, imputation methods based on deep neural networks have shown great success for both deterministic imputation~\citep{cao2018brits,che2018GRUD,luo2018gan} and probabilistic imputation~\citep{fortuin2020gp}. These imputation methods typically utilize autoregressive models to deal with time series.

\begin{figure*}[htbp]
\centering
        \begin{center}
          \includegraphics[width=.89\textwidth]{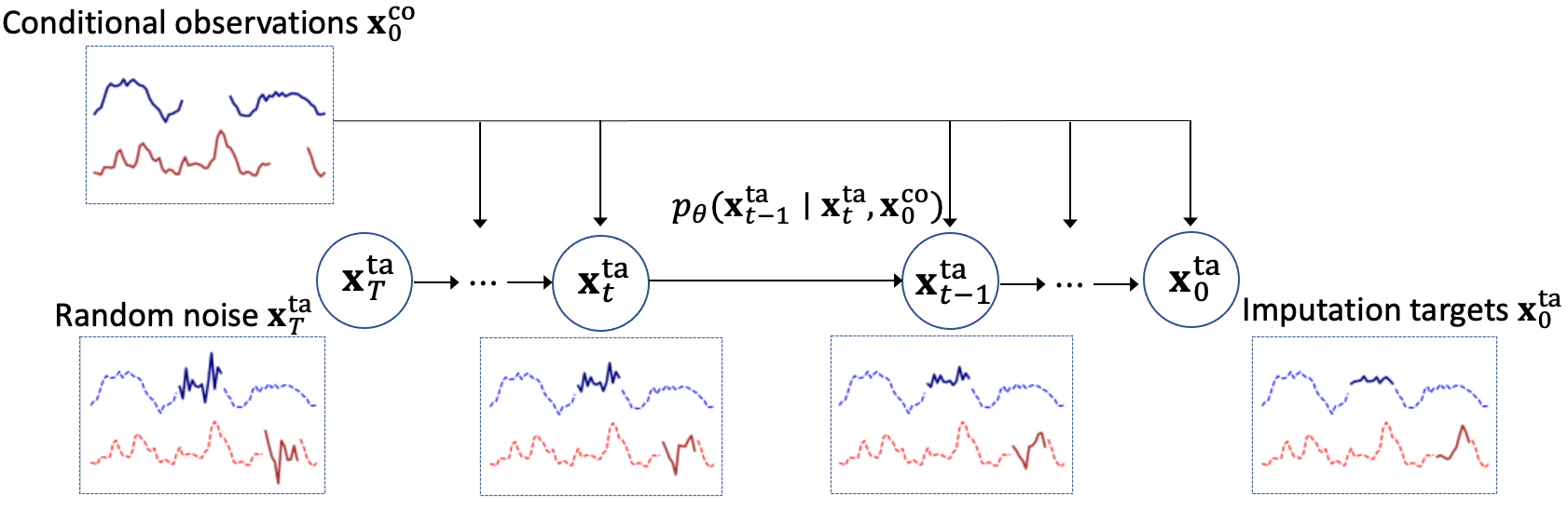}
        \end{center}
    \caption{ The procedure of time series imputation with \pname. The reverse process $p_\theta$ gradually converts random noise into plausible time series, conditioned on observed values $\bmx_0^\rmo$. Dashed lines in each box represent observed values, which are plotted in order to show the relationship with generated imputation and not included in each $\bmx_t^\rmu$. }
    \label{fig:intro_keyvisual}
\end{figure*}

Score-based diffusion models
-- a class of deep generative models and generate samples by gradually converting noise into a plausible data sample through denoising
-- %
have recently achieved state-of-the-art sample quality in many tasks such as image generation~\citep{ho2020denoising,song2021score} and audio synthesis~\citep{kong2020diffwave,chen2021wavegrad}, outperforming counterparts including autoregressive models. Diffusion models 
can also be used to impute missing values by approximating the scores of the posterior distribution obtained from the prior by conditioning on the observed values~\citep{song2021score,kadkhodaie2020denoiseimpute,mittal2021symbolicDiffusion}. While these approximations may work well in practice, they do not correspond to the exact conditional distribution.

In this paper, we propose \pname, a novel probabilistic imputation method that directly learns the conditional distribution with \emph{conditional} score-based diffusion models. Unlike existing score-based approaches, the conditional diffusion model is designed for imputation and can exploit useful information in observed values. We illustrate the procedure of time series imputation with \pname{} in \Figref{fig:intro_keyvisual}. 
We start imputation from random noise on the left of the figure and gradually convert the noise into plausible time series through the reverse process $p_\theta$ of the conditional diffusion model. At each step $t$, the reverse process removes noise from the output of the previous step ($t+1$). %
Unlike existing score-based diffusion models, the reverse process can take observations (on the top left of the figure) as a conditional input, allowing the model to exploit information in the observations for denoising.
We utilize an attention mechanism to capture the temporal and feature dependencies of time series. 

For training the conditional diffusion model, we need observed values (i.e., conditional information) and ground-truth missing values (i.e., imputation targets).  However, in practice we do not know the ground-truth missing values, or training data may not contain missing values at all. Then, inspired by masked language modeling, we develop a self-supervised training method that separates observed values into conditional information and imputation targets. We note that \pname{} is formulated for general imputation tasks, and is not restricted to time series imputation.

Our main contributions are as follows:
\begin{itemize}
    \item We propose conditional score-based diffusion models for probabilistic imputation (\pname), and implement \pname{} for time series imputation. To train the conditional diffusion model, we develop a self-supervised training method. 
    \item We empirically show that \pname{} improves the continuous ranked probability score (CRPS) by 40-65\% over existing probabilistic methods on healthcare and environmental data. Moreover, deterministic imputation with \pname{} decreases the mean absolute error (MAE) by 5-20\% compared to the state-of-the-art methods developed for deterministic imputation. 
    \item We demonstrate that \pname{} can also be applied to time series interpolations and probabilistic forecasting, and is competitive with existing baselines designed for these tasks.
\end{itemize}

\section{Related works}
\label{sec:related}
\paragraph{Time series imputations with deep learning}
Previous studies have shown deep learning models can capture the temporal dependency of time series and give more accurate imputation than statistical methods. 
A popular approach using deep learning is to use RNNs, including LSTMs and GRUs, for sequence modeling~\citep{yoon2018MRNN,che2018GRUD,cao2018brits}. 
Subsequent studies combined RNNs with other methods to improve imputation performance, such as GANs~\citep{luo2018gan,luo2019e2gan,miao2021SSGAN} and self-training~\citep{choi2020rdis}. 
Among them, the combination of RNNs with attention mechanisms is particularly successful for imputation and interpolation of time series~\citep{suo2020glima,shukla2021multi}. 
While these methods focused on deterministic imputation,  GP-VAE~\citep{fortuin2020gp} has been recently developed as a probabilistic imputation method.

\paragraph{Score-based generative models} 
Score-based generative models, including score matching with Langevin dynamics~\citep{song2019generative} and denoising diffusion probabilistic models~\cite{ho2020denoising}, have outperformed existing methods with other deep generative models in many domains, such as images~\citep{song2019generative,ho2020denoising}, audio~\citep{kong2020diffwave,chen2021wavegrad}, and graphs~\citep{niu2020permutationgraph}. 
Most recently, TimeGrad~\citep{rasul2021autoregressive} utilized diffusion probabilistic models for probabilistic time series forecasting. While the method has shown state-of-the-art performance, it cannot be applied to time series imputation due to the use of RNNs to handle past time series.

\section{Background}
\label{sec:background}

\subsection{Multivariate time series imputation}
\label{sec:background:deftime}
We consider $N$ \emph{multivariate} time series with missing values. 
Let us denote the values of each time series as $\bmX = \{x_{1:K,1:L}\} \in \bbR^{K\times L}$ where $K$ is the number of features and $L$ is the length of time series. 
While the length $L$ can be different for each time series, we treat the length of all time series as the same for simplicity, unless otherwise stated. 
We also denote an observation mask as $\bmMtime =\{m_{1:K,1:L}\} \in\{0,1\}^{K\times L}$ where $m_{k,l}=0$ if $x_{k,l}$ is missing, and $m_{k,l}=1$ if $x_{k,l}$ is observed. 
We assume time intervals between two consecutive data entries can be different, 
and define the timestamps of the  time series as $\bms=\{s_{1:L}\} \in \bbR^{L}$. %
In summary, each time series is expressed as $\{\bmX, \bmMtime, \bms \}$. 

Probabilistic time series imputation is the task of estimating the distribution of the missing values of $\bmX$ by exploiting the observed values of $\bmX$. 
We note that this definition of imputation includes other related tasks, such as interpolation, which imputes all features at target time points, and forecasting, which imputes all features at future time points.

\subsection{Denoising diffusion probabilistic models}
\label{sec:background:ddpm}
Let us consider learning a model distribution $p_\theta(\bmx_0)$ that approximates a data distribution $q(\bmx_0)$. Let $\bmx_t$ for $t=1,\ldots,T$ be a sequence of latent variables in the same sample space as $\bmx_0$, which is denoted as $\gX$. 
Diffusion probabilistic models~\citep{sohl2015ddpm} are latent variable models that are composed of two processes: the forward process and the reverse process. %
The forward process is defined by the following Markov chain:
\begin{align}
    q(\bmx_{1:T} \mid \bmx_0) := \prod_{t=1}^{T} q(\bmx_t \mid \bmx_{t-1}) \ \text{where} \ q(\bmx_t \mid \bmx_{t-1}) := \gN\left(\sqrt{1-\beta_t} \bmx_{t-1}, \beta_t \bmI\right) 
    \label{eq:forward}
\end{align}
and $\beta_t$ is a small positive constant that represents a noise level. 
Sampling of $\bmx_t$ has the closed-form written as 
$q(\bmx_t \mid \bmx_0) = \gN (\bmx_t; \sqrt{\alpha_t}\bmx_0, (1-\alpha_t)\bmI)$ where $\hat{\alpha}_t:=1-\beta_t$ and $\alpha_t := \prod_{i=1}^t \hat{\alpha}_i$. Then, $\bmx_t$ can be expressed as $\bmx_t = \sqrt{\alpha_t}\bmx_0 + (1-\alpha_t) \rvepsilon$ where $\rvepsilon \sim \gN(\bmzero,\bmI)$. 
On the other hand, the reverse process denoises $\bmx_t$ to recover $\bmx_0$, and is defined by the following Markov chain:
\begin{align}
\begin{aligned}
     & p_\theta(\bmx_{0:T}) := p(\bmx_T) \prod_{t=1}^{T} p_\theta(\bmx_{t-1} \mid \bmx_t), \quad 
      \bmx_T \sim \gN(\bmzero,\bmI), \\
    &  p_\theta(\bmx_{t-1} \mid \bmx_t) :=   \gN(\bmx_{t-1}; \bmmu_\theta(\bmx_t,t),\sigma_\theta(\bmx_t,t)\bmI).
      \label{eq:reverse}
\end{aligned}
\end{align}
Ho et al.~\citep{ho2020denoising} has recently proposed denoising diffusion probabilistic models (DDPM), which considers the following specific parameterization of $p_\theta(\bmx_{t-1} \mid \bmx_t)$:
\begin{align}
  \bmmu_\theta(\bmx_t,t)=\frac{1}{\alpha_t} \left(\bmx_t-\frac{\beta_t}{\sqrt{1-\alpha_t}} \rvepsilon_\theta (\bmx_t,t) \right), \ 
  \sigma_\theta(\bmx_t,t) = \tilde{\beta}_t^{1/2} \  \mathrm{where} \ \tilde{\beta}_t=\begin{cases}
  \frac{1-\alpha_{t-1}}{1-\alpha_t} \beta_t & t > 1\\
  \beta_1 & t=1
  \end{cases}
  \label{eq:mu}
\end{align}
where $\rvepsilon_\theta$ is a trainable denoising function. We denote $\bmmu_\theta(\bmx_t,t)$ and $\sigma_\theta(\bmx_t,t)$ in \eqref{eq:mu} as $\bmmu^{\textrm{DDPM}}(\bmx_t,t,\rvepsilon_\theta(\bmx_t,t))$ and 
$\sigma^{\textrm{DDPM}}(\bmx_t,t)$, respectively.
The denoising function in \eqref{eq:mu} also corresponds to a rescaled score model for score-based generative models~\citep{song2019generative}. 
Under this parameterization, Ho et al.~\citep{ho2020denoising} have shown that the reverse process can be trained by solving the following optimization problem:
\begin{align}
  \min_{\theta} \gL(\theta) := \min_{\theta} 
  \bb{E}_{\bmx_0 \sim q(\bmx_0), \rvepsilon \sim \gN(\bmzero, \bmI),t}
  || \rvepsilon - \rvepsilon_{\theta}(\bmx_t, t) ||_2^2 \quad \textrm{where} \ \bmx_t= \sqrt{{\alpha}_t}\bmx_0 + (1-{\alpha}_t) \rvepsilon.
  \label{eq:loss}
\end{align}
The denoising function $\rvepsilon_{\theta}$ estimates the noise vector $\rvepsilon$ that was added to its noisy input $\bmx_t$.  This training objective also be viewed as a weighted combination of denoising score matching used for training score-based generative models~\citep{song2019generative,song2020improved,song2021score}.  
Once trained, we can sample $\bmx_0$ from \eqref{eq:reverse}. We provide the details of DDPM in \Appref{append:ddpm}.

\subsection{Imputation with diffusion models}
\label{sec:background:imp}
Here, we focus on general imputation tasks that are not restricted to time series imputation. Let us consider the following imputation problem: given a sample $\bmx_0$ which contains missing values, we generate imputation targets $\bmx_0^\rmu \in \gX^\rmu$ by exploiting conditional observations $\bmx_0^\rmo\in \gX^\rmo$, where $\gX^\rmu$ and $\gX^\rmo$ are a part of the sample space $\gX$ and vary per sample. 
Then, the goal of probabilistic imputation is to estimate the true  conditional data distribution $q({\bmx}_0^\rmu \mid \bmx_0^\rmo)$ with a model distribution $p_\theta({\bmx}_0^\rmu \mid \bmx_0^\rmo)$. 
We typically impute all missing values using all observed values, and set all observed values as $\bmx_0^\rmo$ and all missing values as $\bmx_0^\rmu$, respectively. 
Note that time series imputation in \Secref{sec:background:deftime} can be considered as a special case of this task.

Let us consider modeling $p_\theta({\bmx}_0^\rmu \mid \bmx_0^\rmo)$ with a diffusion model. In the unconditional case, the reverse process $p_\theta(\bmx_{0:T})$ is used to define the final data model $p_\theta(\bmx_0)$. Then, a natural approach is to extend the reverse process in \eqref{eq:reverse} to a conditional one:
\begin{align}
\begin{aligned}
      &p_\theta(\bmx_{0:T}^\rmu \mid \bmx_0^\rmo) := p(\bmx_T^\rmu) \prod_{t=1}^{T} p_\theta(\bmx_{t-1}^\rmu \mid \bmx_t^\rmu,\bmx_0^\rmo), \quad
      \bmx_T^\rmu \sim \gN(\bmzero,\bmI), \\
     &p_\theta(\bmx_{t-1}^\rmu \mid \bmx_t^\rmu,\bmx_0^\rmo) := \gN(\bmx_{t-1}^\rmu; \bmmu_\theta(\bmx_t^\rmu,t \mid \bmx_0^\rmo),\sigma_\theta(\bmx^\rmu_t,t \mid \bmx_0^\rmo)\bmI ).
\label{eq:reverse_cond_part}
\end{aligned}
\end{align}
However, existing diffusion models are generally designed for data generation and do not take conditional observations $\bmx_0^\rmo$ as inputs. 
To utilize diffusion models for imputation, previous studies~\citep{song2021score,kadkhodaie2020denoiseimpute,mittal2021symbolicDiffusion} approximated the conditional reverse process $p_\theta(\bmx_{t-1}^\rmu \mid \bmx_t^\rmu,\bmx_0^\rmo)$ with the reverse process in \eqref{eq:reverse}. 
With this approximation, in the reverse process they add noise to both the target and the conditional observations $\bmx_0^\rmo$.
While this approach can impute missing values, 
the added noise can harm useful information 
in the observations. This suggests that modeling  $p_\theta(\bmx_{t-1}^\rmu \mid \bmx_t^\rmu,\bmx_0^\rmo)$ without approximations can improve the imputation quality. Hereafter, 
we call the model defined in \Secref{sec:background:ddpm} as the unconditional diffusion model.

\section{Conditional score-based diffusion model for imputation (\pname)}
\label{sec:model}

In this section, we propose \pname, a novel imputation method based on a conditional score-based diffusion model. The conditional diffusion model allows us to exploit useful information in observed values for accurate imputation. We provide the reverse process of the conditional diffusion model, and then develop a self-supervised training method. %
We note that \pname{} is not restricted to time series. 

\subsection{Imputation with \pname}
\label{sec:model:conditional}
We focus on the conditional diffusion model with the reverse process in \eqref{eq:reverse_cond_part} and aim to model the conditional distribution $p(\bmx^\rmu_{t-1} \mid \bmx^\rmu_t ,\bmx^\rmo_0)$ without approximations. Specifically, we extend the parameterization of DDPM in \eqref{eq:mu} to the conditional case. We define a conditional denoising function $\rvepsilon_\theta: (\gX^\rmu\times \bbR \mid \gX^\rmo)\rightarrow \gX^\rmu$, which takes conditional observations $\bmx_0^\rmo$ as inputs. 
Then, we consider the following parameterization with $\rvepsilon_\theta$:
\begin{align}
\begin{aligned}
\bmmu_\theta(\bmx_t^\rmu,t \mid \bmx_0^\rmo) =  \bmmu^{\textrm{DDPM}}(\bmx_t^\rmu,t,\rvepsilon_\theta(\bmx_t^\rmu,t \mid \bmx_0^\rmo)), \quad
\sigma_\theta(\bmx^\rmu_t,t \mid \bmx_0^\rmo) =  \sigma^{\textrm{DDPM}}(\bmx_t^\rmu,t) 
\label{eq:reverse_cond}
\end{aligned}
\end{align}
where $\bmmu^{\textrm{DDPM}}$ and $\sigma^{\textrm{DDPM}}$ are the functions defined in \Secref{sec:background:ddpm}.
Given the function $\rvepsilon_\theta$ and data $\bmx_0$, we can sample $\bmx_0^\rmu$ using the reverse process in \eqref{eq:reverse_cond_part} and \eqref{eq:reverse_cond}.
For the sampling, we set all observed values of $\bmx_0$ as conditional observations $\bmx_0^\rmo$ and all missing values as imputation targets $\bmx_0^\rmu$. Note that the conditional model is reduced to the unconditional one under no conditional observations and can also be used for data generation.

\begin{figure*}[htbp]
\centering
        \begin{center}
          \includegraphics[width=.8\textwidth]{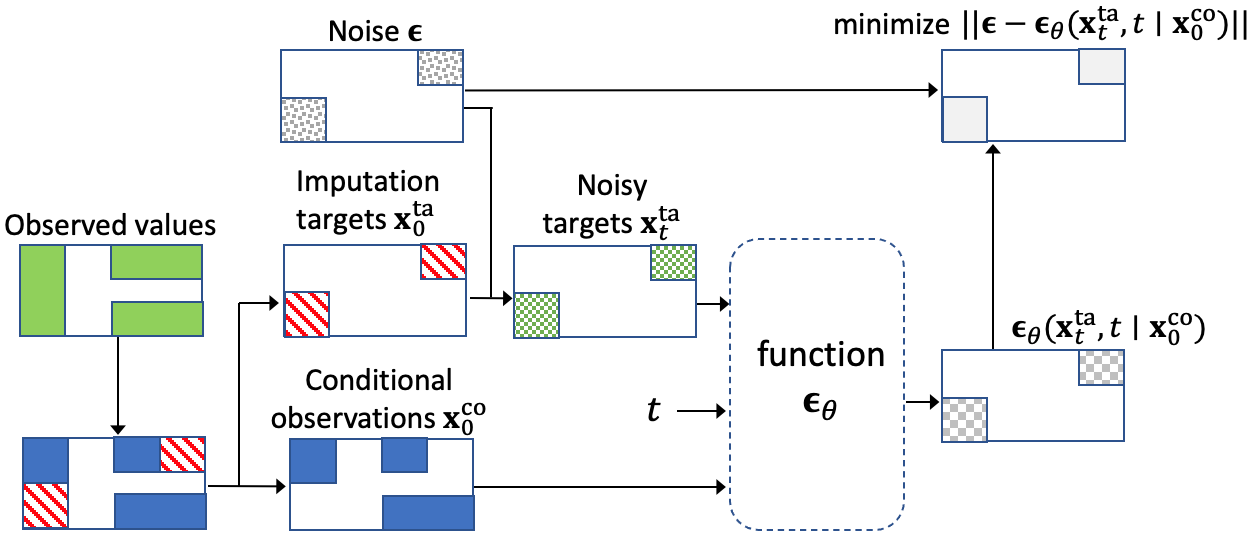}
        \end{center}
    \caption{The self-supervised training procedure of \pname{}. On the middle left rectangle, the green and white areas represent observed and missing values, respectively. The observed values are separated into red imputation targets $\bmx_0^\rmu$ and blue conditional observations $\bmx_0^\rmo$, and used for 
    training of $\rvepsilon_\theta$. The colored areas in each rectangle mean the existence of values. }%
    \label{fig:training}
\end{figure*}

\begin{table*}[htb]
\caption{Imputation targets $\bmx_0^\rmu$ and conditional observations $\bmx_0^\rmo$ for \pname{} at training and sampling.}
\begin{center}
\setlength{\tabcolsep}{5pt}
\begin{tabular}{lcc}
\toprule
& imputation targets  $\bmx_0^\rmu$  & conditional observations $\bmx_0^\rmo$ \\ \midrule
sampling (imputation) & all missing values & all observed values  \\ \midrule
training & 
\begin{tabular}{c}
a subset of the observed values \\
(sampled by a target choice strategy)
\end{tabular}
& 
\begin{tabular}{c}
the remaining \\
 observed values 
\end{tabular}
\\ \bottomrule
\end{tabular}		
\end{center}
\label{tab:summaryinput}
\end{table*}

\subsection{Training of \pname}
\label{sec:model:training}
Since \eqref{eq:reverse_cond} uses the same parameterization as \eqref{eq:mu} and the difference between \eqref{eq:mu} and \eqref{eq:reverse_cond} is only the form of $\rvepsilon_\theta$, we can follow the training procedure for the unconditional model in \Secref{sec:background:ddpm}. 
Namely, given conditional observations $\bmx_0^\rmo$ and imputation targets $\bmx_0^\rmu$, we sample noisy targets $\bmx_t^\rmu = \sqrt{{\alpha}_t}\bmx_0^\rmu +  (1-{\alpha}_t) \rvepsilon$, and train $\rvepsilon_\theta$ by minimizing the following loss function:
\begin{align}
  \min_{\theta} \gL(\theta) := \min_{\theta} 
  \bb{E}_{\bmx_0 \sim q(\bmx_0), \rvepsilon \sim \gN(\bmzero, \bmI),t}
  || (\rvepsilon - \rvepsilon_\theta (\bmx_t^\rmu,t \mid \bmx_0^\rmo)) ||_2^2
  \label{eq:loss_cond}
\end{align}
where the dimension of $\rvepsilon$ corresponds to that of the imputation targets $\bmx_0^\rmu$. 

However, this training procedure has an issue. Since we do not know the ground-truth missing values in practice, it is not clear how to select $\bmx_0^\rmo$ and $\bmx_0^\rmu$ from a training sample $\bmx_0$. To address this issue, we develop a self-supervised learning method inspired by masked language modeling~\citep{devlin-etal-2019-bert}. We illustrate the training procedure in \Figref{fig:training}. Given a sample $\bmx_0$, we separate observed values of $\bmx_0$ into two parts, and set one of them as imputation targets $\bmx_0^\rmu$ and the other as 
conditional observations $\bmx_0^\rmo$. 
We choose the targets $\bmx_0^\rmu$ through a target choice strategy, which is discussed in \Secref{sec:model:targetchoice}. Then, we sample noisy targets $\bmx_t^\rmu$ and train $\rvepsilon_\theta$ by solving \eqref{eq:loss_cond}.
We summarize how we set $\bmx_0^\rmo$ and $\bmx_0^\rmu$ for training and sampling in \Tabref{tab:summaryinput}. We also provide the algorithm of training and sampling in \Appref{append:algorithms:conditional}.

\subsection{Choice of imputation targets in self-supervised learning}
\label{sec:model:targetchoice}
In the proposed self-supervised learning, the choice of imputation targets is important. We provide four target choice strategies depending on what is known about the missing patterns in the test dataset. We describe the algorithm for these strategies in \Appref{append:algorithms:knownmask}. 

(1) {\it Random} strategy %
: this strategy is used when we do not know about missing patterns, and randomly chooses a certain percentage of observed values as imputation targets. The percentage is sampled from $[0\%, 100\%]$ to adapt to various missing ratios in the test dataset.

(2) {\it Historical} strategy: this strategy exploits missing patterns in the training dataset. Given a training sample $\bmx_0$, we randomly draw another sample $\tilde{\bmx}_0$ from the training dataset. Then, we set the intersection of the observed indices of $\bmx_0$ and the missing indices of $\tilde{\bmx}_0$ as imputation targets. 
The motivation of this strategy comes from structured missing patterns in the real world. For example, missing values often appear consecutively in time series data. When missing patterns in the training and test dataset are highly correlated, this strategy helps the model learn a good conditional distribution.

(3) {\it Mix} strategy: this strategy is the mix of the above two strategies. The historical strategy may lead to overfitting to missing patterns in the training dataset. The {\it Mix} strategy can benefit from generalization by the random strategy and structured missing patterns by the historical strategy.

(4) {\it Test} pattern strategy: when we know the missing patterns in the test dataset, we just set the patterns as imputation targets. For example, this strategy is used for time series forecasting, since the missing patterns in the test dataset are fixed to given future time points.

\section{Implementation of \pname{} for time series imputation}
\label{sec:main:time}

\begin{figure*}[htbp]
\centering
        \begin{center}
          \includegraphics[width=.95\textwidth]{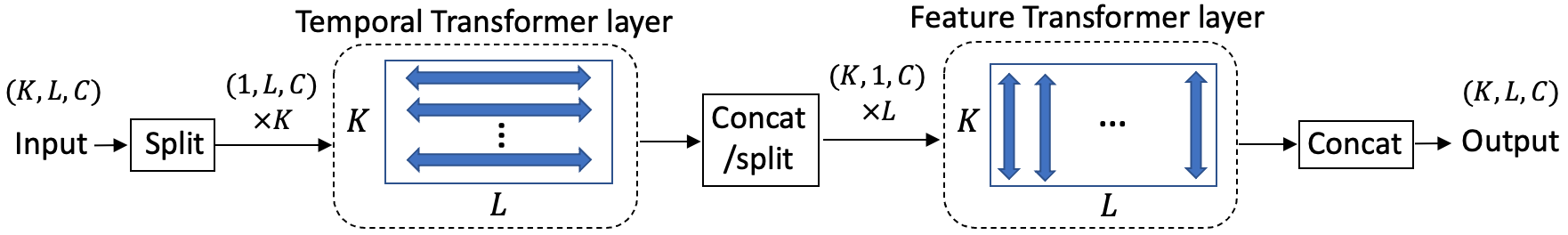}
        \end{center}
    \caption{The architecture of 2D attention. Given a tensor with $K$ features, $L$ length, and $C$ channels, the temporal Transformer layer takes tensors with $(1,L,C)$ shape as inputs and learns temporal dependency. The feature Transformer layer takes tensors with $(K,1,C)$ shape as inputs and learns feature dependency. The output shape of each layer is the same as the input shape. }
    \label{fig:2Dattention}
\end{figure*}

\label{sec:main:architecture} 
In this section, we implement \pname{} for time series imputation. For the implementation, we need the inputs and the architecture of $\rvepsilon_\theta$. 

First, we describe how we process time series data as inputs for \pname. 
As defined in \Secref{sec:background:deftime}, a time series is denoted as \{\bmX, \bmMtime, \bms \}, and the sample space $\gX$ of $\bmX$ is $\bbR^{K\times L}$. 
We want to handle $\bmX$ in the sample space $\bbR^{K\times L}$ for learning dependencies in a time series using a neural network, but the conditional denoising function $\rvepsilon_\theta$ takes inputs $\bmx^\rmu_t$ and $\bmx_0^\rmo$ in varying sample spaces that are a part of $\gX$ 
as shown in white areas of $\bmx^\rmu_t$ and $\bmx_0^\rmo$ in \Figref{fig:training}.  
To address this issue, we adjust the conditional denoising function $\rvepsilon_\theta$ to inputs in the fixed sample space $\bbR^{K\times L}$. 
Concretely, we fix the shape of the inputs $\bmx^\rmu_t$ and $\bmx_0^\rmo$ to $(K\times L)$ by applying zero padding to $\bmx^\rmu_t$ and $\bmx_0^\rmo$. In other words, we set zero values to white areas for $\bmx^\rmu_t$ and $\bmx_0^\rmo$ in \Figref{fig:training}.
To indicate which indices are padded, we introduce the conditional mask $\bmM^\rmo \in \{0,1\}^{K\times L}$ as an additional input to $\rvepsilon_\theta$, which corresponds to $\bmx_0^\rmo$ and takes value 1 for indices of conditional observations. 
For ease of handling, we also fix the output shape in the sample space $\bbR^{K\times L}$ by applying zero padding. Then, the conditional denoising function $\rvepsilon_\theta(\bmx_t^\rmu,t \mid \bmx_0^\rmo, \bmM^\rmo)$ can be written as $\rvepsilon_\theta:(\bbR^{K\times L}\times \bbR \mid \bbR^{K\times L}\times \{0,1\}^{K\times L})\rightarrow \bbR^{K\times L}$.
We discuss the effect of this adjustment on training and sampling in \Appref{append:model_extend}.

Under the adjustment, we set conditional observations $\bmx_0^\rmo$ and imputation targets $\bmx_0^\rmu$ for time series imputation by following \Tabref{tab:summaryinput}. At sampling time, since conditional observations $\bmx_0^\rmo$ are all observed values, we set $\bmM^\rmo=\bmMtime$ and $\bmx_0^\rmo=\bmM^\rmo \odot \bmX$ where $\odot$ represents element-wise products. For training, we sample $\bmx_0^\rmu$ and $\bmx_0^\rmo$ through a target choice strategy, and set the indices of $\bmx_0^\rmo$ as $\bmM^\rmo$. Then, $\bmx_0^\rmo$ is written as $\bmx_0^\rmo=\bmM^\rmo \odot \bmX$ and $\bmx_0^\rmu$ is obtained as $\bmx_0^\rmu=(\bmMtime-\bmM^\rmo) \odot \bmX$.

Next, we describe the architecture of $\rvepsilon_\theta$. We adopt the architecture in DiffWave~\citep{kong2020diffwave} as the base, which is composed of multiple residual layers with residual channel $C$.
We refine this architecture for time series imputation. We set the diffusion step $T=50$. 
We discuss the main differences from DiffWave (see \Appref{append:settings:diffusion_model} for the whole architecture and details).
\paragraph{Attention mechanism} %
To capture temporal and feature dependencies of multivariate time series, we utilize a two dimensional attention mechanism in each residual layer instead of a convolution architecture.
As shown in \Figref{fig:2Dattention}, we introduce temporal Transformer layer and a feature Transformer layer, which are 1-layer Transformer encoders. 
The temporal Transformer layer takes tensors for each feature as inputs to learn temporal dependency, whereas the feature Transformer layer takes tensors for each time point as inputs to learn temporal dependency.

Note that while the length $L$ can be different for each time series as mentioned in \Secref{sec:background:deftime}, the attention mechanism allows the model to handle various lengths. For batch training, we apply zero padding to each sequence so that the lengths of the sequences are the same.

\paragraph{Side information} 
In addition to the arguments of $\rvepsilon_\theta$, we provide some side information as additional inputs to the model. First, we use time embedding of $\bms=\{s_{1:L}\}$ to learn the temporal dependency.
Following previous studies~\citep{vaswani2017attention,zuo2020transformer}, we use 128-dimensions temporal embedding. Second, we exploit categorical feature embedding for $K$ features, where the dimension is 16. %

\section{Experimental results}
\label{sec:experiments}

In this section, we demonstrate the effectiveness of \pname{} for time series imputation. Since \pname{} can be applied to other related tasks such as interpolation and forecasting, we also evaluate \pname{} for these tasks to show the flexibility of \pname. Due to the page limitation, we provide the detailed setup for experiments including train/validation/test splits and hyperparameters in \Appref{append:settings:experiments}.

\subsection{Time series imputation}

\label{sec:experiments:imp}
\paragraph{Dataset and experiment settings}
We run experiments for two datasets. The first one is the healthcare dataset in PhysioNet Challenge 2012~\citep{silva2012physionet}, which consists of 4000 clinical time series with 35 variables for 48 hours from intensive care unit (ICU). Following previous studies~\citep{cao2018brits,che2018GRUD}, we process the dataset to hourly time series with 48 time steps. The processed dataset contains around 80\% missing values.
Since the dataset has no ground-truth, we randomly choose 10/50/90\% of observed values as ground-truth on the test data.

The second one is the air quality dataset~\citep{yi2016st}.
Following previous studies~\citep{cao2018brits,suo2020glima}, we use hourly sampled PM2.5 measurements from 36 stations in Beijing for 12 months and set 36 consecutive time steps as one time series. There are around 13\% missing values and the missing patterns are not random. The dataset contains artificial ground-truth, whose missing patterns are also structured.

For both dataset, we run each experiment five times. As the target choice strategy for training, we adopt the random strategy for the healthcare dataset and the mix of the random and historical strategy for the air quality dataset, based on the missing patterns of each dataset.

\paragraph{Results of probabilistic imputation}
\pname{} is compared with three baselines. 
1) Multitask GP~\citep{multitaskgp}: the method learns the covariance between timepoints and features simultaneously. 
2) GP-VAE~\citep{fortuin2020gp}: the method showed the state-of-the-art results for probabilistic imputation. %
3) V-RIN~\citep{mulyadi2021uncertainty}: a deterministic imputation method that uses the uncertainty quantified by VAE to improve imputation. For V-RIN, we regard the quantified uncertainty as probabilistic imputation.  
In addition, we compare \pname{} with imputation using the unconditional diffusion model in order to show the effectiveness of the conditional one (see \Appref{append:uncond} for 
training and imputation with 
the unconditional diffusion model). 

We first show quantitative results. We adopt the continuous ranked probability score (CRPS)~\citep{matheson1976CRPS} as the metric, which is freuquently used for evaluating probabilistic time series forecasting and measures the compatibility of an estimated probability distribution with an observation. 
We generate 100 samples to approximate the probability distribution over missing values and report the normalized average of CRPS for all missing values following previous studies~\citep{salinas2019high} (see \Appref{append:CRPS} for details of the computation).

\begin{table*}[htb]
\caption{Comparing CRPS for probabilistic imputation baselines and \pname{} (lower is better). 
We report the mean and the standard error of CRPS for five trials.
}
\begin{center}
\begin{tabular}{lcccc}
\toprule
& \multicolumn{3}{c} { healthcare } & air quality \\
\cmidrule{2-4} & $10 \%$ missing & $50 \%$ missing& $90 \%$ missing& \\
\midrule 
Multitask GP~\citep{multitaskgp} & $0.489(0.005)$ & $0.581(0.003)$ & $0.942(0.010)$ & $0.301(0.003)$ \\
GP-VAE~\citep{fortuin2020gp} & $0.574(0.003)$ & $0.774(0.004)$ & $0.998(0.001)$ & $0.397(0.009)$ \\
 V-RIN~\citep{mulyadi2021uncertainty} & $0.808(0.008)$ & $0.831(0.005)$ & $0.922(0.003)$ & $0.526(0.025)$ \\ 
 unconditional & $0.360(0.007)$ &$0.458(0.008)$ & $0.671(0.007)$ & $0.135(0.001)$\\
 {\bf \pname} (proposed) & ${\bf 0.238(0.001)}$ & ${\bf 0.330(0.002)}$ & ${\bf 0.522(0.002)}$ & ${\bf 0.108(0.001)}$ \\
\bottomrule		
\end{tabular}		
\end{center}

\label{tab:result_prob}
\end{table*}

\begin{figure}[htbp]
\centering

    \begin{tabular}{cc}
      \begin{minipage}{0.4\hsize} %
        \begin{center}
          \includegraphics[width=1.0\textwidth]{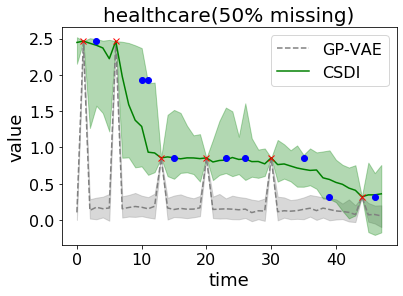}
        \end{center}
      \end{minipage}
        &
      \begin{minipage}{0.4\hsize}
        \begin{center}
          \includegraphics[width=1.0\textwidth]{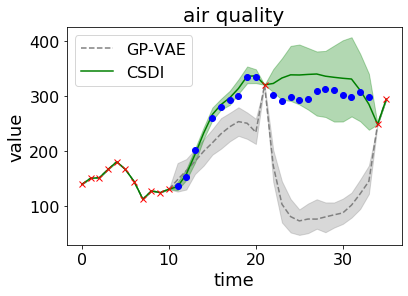}
        \end{center}
      \end{minipage}
    \end{tabular}
    \caption{Examples of probabilistic time series imputation for the healthcare dataset with 50\% missing (left) and the air quality dataset (right). The red crosses show the observed values and the blue circles show the ground-truth imputation targets. For each method, median values of imputations are shown as the line and 5\% and 95\% quantiles are shown as the shade. } %
    \label{fig:exp_example_imputation}
\end{figure}

\Tabref{tab:result_prob} represents CRPS for each method. \pname{} reduces CRPS by 40-65\% compared to the existing baselines for both datasets. This indicates that \pname{} generates more realistic distributions than other methods.  
We also observe that the imputation with \pname{} outperforms that with the unconditional model. This suggests \pname{} benefits from explicitly modeling the conditional distribution.

We provide imputation examples in \Figref{fig:exp_example_imputation}. For the air quality dataset, \pname{} (green solid line) provides accurate imputations with high confidence, while those by GP-VAE (gray dashed line) are far from ground-truth. 
\pname{} also gives reasonable imputations for the healthcare dataset.
These results indicate that \pname{} exploits temporal and feature dependencies to provide accurate imputations.
We give more examples in \Appref{append:examples}.

\begin{table*}[htb]
\caption{Comparing MAE for deterministic imputation methods and \pname. We report the mean and the standard error for five trials. The asterisks mean the results of the method are cited from the original paper.%
}
\begin{center}
\begin{tabular}{lcccc} 
\toprule
& \multicolumn{3}{c} { healthcare } & air quality \\
\cmidrule{2-4} & $10 \%$ missing & $50 \%$ missing& $90 \%$ missing& \\
\midrule V-RIN~\citep{mulyadi2021uncertainty}& $0.271(0.001)$ & $0.365(0.002)$ & $0.606(0.006)$ & $25.4(0.62)$ \\
 BRITS~\citep{cao2018brits} & $0.284(0.001)$ & $0.368(0.002)$ & $0.517(0.002)$ & $14.11(0.26)$ \\
 BRITS~\citep{cao2018brits} (*) & $0.278$ & $-$ & $-$ & $11.56$ \\
 GLIMA~\citep{suo2020glima} (*) & $0.265$ & $-$ & $-$ & $10.54$ \\ 
 RDIS~\citep{choi2020rdis} & $0.319(0.002)$ & $0.419(0.002)$ &	$0.631(0.002)$& $22.11(0.35)$\\ 
 unconditional & $0.326(0.008)$ & $0.417(0.010)$ & $0.625(0.010)$ & $12.13(0.07)$ \\
 {\bf \pname} (proposed) & ${\bf 0.217(0.001)}$ &  ${\bf 0.301(0.002)}$ & ${\bf 0.481(0.003)}$ & ${\bf 9.60(0.04)}$ \\
\bottomrule
\end{tabular}	
\end{center}
\label{tab:result_det}
\end{table*}

\paragraph{Results of deterministic imputation}
We demonstrate that \pname{} also provides accurate deterministic imputations, which are obtained as the median of 100 generated samples. 
We compare \pname{} with four baselines developed for deterministic imputation including GLIMA~\citep{suo2020glima}, which combined recurrent imputations with an attention mechanism to capture temporal and feature dependencies and showed the state-of-the-art performance. These methods are based on autoregressive models. We use the original implementations except RDIS.

We evaluate each method by the mean absolute error (MAE). In \Tabref{tab:result_det}, \pname{} improves MAE by 5-20\% compared to the baselines. This suggests that the conditional diffusion model is effective to learn temporal and feature dependencies for imputation. For the healthcare dataset, the gap between the baselines and \pname{} is particularly significant when the missing ratio is small, because more observed values help \pname{} capture dependencies. 

\begin{table*}[htb]
\caption{Comparing the state-of-the-art interpolation methods with \pname{} for the healthcare dataset. We report the mean and the standard error of CRPS for five trials.}
\begin{center}
\begin{tabular}{lccc}
\toprule
& $10 \%$ missing & $50 \%$ missing& $90 \%$ missing \\ \midrule
 Latent ODE~\citep{rubanova2019latentODE} & $0.700(0.002)$ & $0.676(0.003)$ & $0.761(0.010)$ \\
 mTANs~\citep{shukla2021multi} & $0.526(0.004)$ & $0.567(0.003)$ & $0.689(0.015)$ \\
 {\bf \pname} (proposed)&  ${\bf 0.380(0.002)}$ & ${\bf 0.418(0.001)}$ & ${\bf 0.556(0.003)}$ \\
\bottomrule
\end{tabular}
\end{center}
\label{tab:result_interp}
\end{table*}

\subsection{Interpolation of irregularly sampled time series}
\label{sec:experiments:interp}

\paragraph{Dataset and experiment settings} 
We use the same healthcare dataset as the previous section, but process the dataset as irregularly sampled time series, following previous studies~\citep{shukla2021multi,rubanova2019latentODE}.
Since the dataset has no ground-truth, we randomly choose 10/50/90\% of {\it time} points and use observed values at these time points as ground-truth on the test data. As the target choice strategy for training, we adopt the random strategy, which is adjusted for interpolation so that some {\it time} points are sampled.

\paragraph{Results}
We compare \pname{} with two baselines including  mTANs~\citep{shukla2021multi}, which utilized an attention mechanism and showed state-of-the-art results for the interpolation of irregularly sampled time series. We generate 100 samples to approximate the probability distribution as with the previous section.
The result is shown in \Tabref{tab:result_interp}. \pname{} outperforms the baselines for all cases. 

\begin{table*}[htb]
\caption{Comparing probabilistic forecasting methods with \pname. We report the mean and the standard error of CRPS-sum for three trials. The baseline results are cited from the original paper. 'TransMAF' is the abbreviation for 'Transformer MAF'.}
\label{tab:result_forecast}

\begin{center}
\setlength{\tabcolsep}{2.5pt}
\begin{tabular}{lccccc}
\toprule & solar & electricity & traffic & taxi & wiki \\
\midrule GP-copula~\citep{salinas2019high} & $0.337(0.024)$ & $0.024(0.002)$ & $0.078(0.002)$ & $0.208(0.183)$ & $0.086(0.004)$ \\
 TransMAF~\citep{rasul2020multi} & $0.301(0.014)$ & $0.021(0.000)$ & $0.056(0.001)$ & $0.179(0.002)$ & $0.063(0.003)$ \\
 TLAE~\citep{nguyen2021temporal} & ${\bf 0.124(0.033)}$ & $0.040(0.002)$ & $0.069(0.001)$ & $0.130(0.006)$ & $0.241(0.001)$ \\
 TimeGrad~\citep{rasul2021autoregressive} & $0.287(0.020)$ & $0.021(0.001)$ & $0.044(0.006)$ & ${\bf 0.114(0.020)}$ & $ 0.049(0.002)$ \\ 
 {\bf \pname} (proposed) & $0.298(0.004)$ & ${\bf 0.017(0.000)}$ & ${\bf 0.020(0.001)}$ & ${ 0.123(0.003)}$ & ${\bf 0.047(0.003)}$ \\				
\bottomrule
\end{tabular}
\end{center}
\end{table*}

\subsection{Time series Forecasting}
\label{sec:experiments:forecast}
\paragraph{Dataset and Experiment settings}
We use five datasets that are commonly used for evaluating probabilistic time series forecasting. Each dataset is composed of around 100 to 2000 features.  
We predict all features at future time steps using past time series. We use the same prediction steps as previous studies~\citep{salinas2019high,nguyen2021temporal}.
For the target choice strategy, we adopt the {\it Test} pattern strategy.

\paragraph{Results}
We compare \pname{} with four baselines. Specifically, TimeGrad~\citep{rasul2021autoregressive} combined the diffusion model with a RNN-based encoder. We evaluate each method for CRPS-sum, which is CRPS for the distribution of the sum of all time series across $K$ features and accounts for joint effect (see \Appref{append:CRPS} for details).

In \Tabref{tab:result_forecast}, \pname{} outperforms the baselines for electricity and traffic datasets, 
and is competitive with the baselines as a whole. 
The advantage of \pname{} over baselines for forecasting is smaller than that for imputation in \Secref{sec:experiments:imp}.
We hypothesize it is because the datasets for forecasting seldom contains missing values and are suitable for existing encoders including RNNs. For imputation, it is relatively difficult for RNNs to handle time series due to missing values.

\section{Conclusion}
\label{sec:conclusion}

In this paper, we have proposed \pname, a novel approach to impute multivariate time series with conditional diffusion models. We have shown that \pname{} outperforms the existing probabilistic and deterministic imputation methods.

There are some interesting directions for future work. 
One direction is to improve the computation efficiency. While diffusion models generate plausible samples, sampling is generally slower than other generative models. To mitigate the issue, several recent studies leverage an ODE solver to accelerate the sampling procedure~\citep{song2021score,song2020DDIM,kong2020diffwave}. %
Combining our method with these approaches would likely improve the sampling efficiency.

Another direction is to extend \pname{} to downstream tasks such as classifications. Many previous studies have shown that accurate imputation improves the performance on downstream tasks~\citep{cao2018brits,luo2019e2gan,shukla2021multi}.
Since conditional diffusion models can learn temporal and feature dependencies with uncertainty, joint training of imputations and downstream tasks using conditional diffusion models would be helpful to improve the performance of the downstream tasks. 

Finally, although our focus was on time series, it would be interesting to explore \pname{} as imputation technique on other modalities.

\section*{Acknowledgements and Disclosure of Funding}
This research was supported by NSF(\#1651565, \#1522054, \#1733686), ONR (N000141912145), AFOSR (FA95501910024), ARO (W911NF-21-1-0125) and Sloan Fellowship.

\bibliography{arxiv_main}
\bibliographystyle{unsrt}


\newpage

\appendix

\section{Details of denoising diffusion probabilistic models}
\label{append:ddpm}
In this section, we describe the details of denoising diffusion probabilistic models in \Secref{sec:background:ddpm}.

Diffusion probabilistic models~\citep{sohl2015ddpm} are latent variable models that are
composed of two processes: the forward process and the reverse process. 
The forward process and the reverse process are defined by \eqref{eq:forward} and~\ref{eq:reverse}, respectively.
Then, the parameters $\theta$ are learned by maximizing variational lower bound (ELBO) of likelihood $p_\theta(\bmx_{0:T})$:
\begin{align}
    \bbE_{q(\bmx_0)}[\log p_\theta(\bmx_0)]\geq \bbE_{q(\bmx_0,\bmx_1,\ldots,\bmx_T)}[\log p_\theta(\bmx_{0:T})-\log q(\bmx_{1:T}\mid \bmx_0)] := \textrm{ELBO}.
      \label{eq:app:elbo}
\end{align}
To analyse this ELBO, Ho et al.~\citep{ho2020denoising} proposed denoising diffusion probabilistic models (DDPM), which considered the parameterization given by \eqref{eq:mu}.
Under the parameterization, Ho et al.~\citep{ho2020denoising} showed ELBO satisfies the following equation:
\begin{align}
  -\textrm{ELBO} = c + \sum_{t=1}^{T} \kappa_t \bb{E}_{\bmx_0 \sim q(\bmx_0), \rvepsilon \sim \gN(\bmzero, \bmI)}  || \rvepsilon - \rvepsilon_{\theta}(\sqrt{\alpha_t}\bmx_0 + (1-\alpha_t) \rvepsilon, t) ||_2^2
  \label{eq:app:ELBO}
\end{align}
where $c$ is a constant and $\{\kappa_{1:T}\}$ are positive coefficients depending on $\alpha_{1:T}$ and $\beta_{1:T}$. The diffusion process can be trained by minimizing \eqref{eq:app:ELBO}. In addition, Ho et al.~\citep{ho2020denoising} found that minimizing the following unweighted version of ELBO leads to good sample quality:
\begin{align}
  \min_{\theta} L(\theta) := \min_{\theta} 
  \bb{E}_{\bmx_0 \sim q(\bmx_0), \rvepsilon \sim \gN(\bmzero, \bmI),t}
  || \rvepsilon - \rvepsilon_{\theta}(\sqrt{\alpha_t}\bmx_0 + (1-\alpha_t) \rvepsilon, t) ||_2^2.
  \label{eq:app_loss}
\end{align}
The function $\rvepsilon_{\theta}$ estimates noise $\rvepsilon$ in the noisy input. Once trained, we can sample $\bmx_0$ from \eqref{eq:reverse}.

\section{Algorithms}
\label{append:algorithms}
\subsection{Algorithm for training and sampling of \pname}
\label{append:algorithms:conditional}
We provide the training procedure of \pname{} in Algorithm~\ref{alg:training} and the imputation (sampling) procedure with \pname{} in Algorithm~\ref{alg:sampling}, which are described in \Secref{sec:model}.

\begin{algorithm}[htbp]
   \caption{Training of \pname}
   \label{alg:training}
\begin{algorithmic}[1]
   \STATE {\bfseries Input:} distribution of training data $q(\bmx_0)$, a target choice strategy $\gT$, the number of iteration $N_{\rm iter}$, the sequence of noise levels $\{\alpha_t\}$
   \STATE {\bfseries Output:} Trained denoising function $\rvepsilon_\theta$
   \FOR{$i=1$ {\bfseries to} $N_{\rm iter}$}
   \STATE $t \sim \textrm{Uniform}(\{1,\ldots,T\})$, $\bmx_0 \sim q(\bmx_0)$ 
   \STATE Separate observed values of $\bmx_0$ into conditional information $\bmx_0^\rmo$ and imputation targets $\bmx_0^\rmu$ by the target choice strategy $\gT$
   \STATE $\rvepsilon \sim \gN(\bmzero,\bmI)$ where the dimension of $\rvepsilon$ corresponds to $\bmx_0^\rmu$
   \STATE Calculate noisy targets $\bmx_t^\rmu = \sqrt{{\alpha}_t}\bmx_0^\rmu +  (1-{\alpha}_t) \rvepsilon$
   \STATE Take gradient step on $\nabla_\theta || (\rvepsilon - \rvepsilon_\theta (\bmx_t^\rmu,t \mid \bmx_0^\rmo)) ||_2^2$ according to \eqref{eq:loss_cond}
   \ENDFOR
\end{algorithmic}
\end{algorithm}

\begin{algorithm}[htbp]
   \caption{Imputation (Sampling) with \pname}
   \label{alg:sampling}
\begin{algorithmic}[1]
   \STATE {\bfseries Input:} a data sample $\bmx_0$, trained denoising function $\rvepsilon_\theta$ 
   \STATE {\bfseries Output:} Imputed missing values $\bmx_0^\rmu$
   \STATE Denote observed values of $\bmx_0$ as $\bmx_0^\rmo$
   \STATE $\bmx_T^\rmu \sim \gN(\bmzero,\bmI)$ where the dimension of $\bmx_T^\rmu$ corresponds to the missing indices of $\bmx_0$
   \FOR{$t=T$ {\bfseries to} $1$}
   \STATE Sample $\bmx_{t-1}^\rmu$ using \eqref{eq:reverse_cond_part} and \eqref{eq:reverse_cond}
   \ENDFOR
\end{algorithmic}
\end{algorithm}

\begin{algorithm}[htbp]
   \caption{Target choice with the random strategy}
   \label{alg:randomchoice}
\begin{algorithmic}[1]
   \STATE {\bfseries Input:} a training sample $\bmx_0$
   \STATE {\bfseries Output:} conditional information $\bmx_0^\rmo$, imputation targets $\bmx_0^\rmu$
   \STATE Draw target ratio $r \sim \textrm{Uniform}(0,100)$
   \STATE Randomly choose $r\%$ of the observed values of $\bmx_0$ and denote the chosen observations as $\bmx_0^\rmu$, and denote the remaining  observations as $\bmx_0^\rmo$ 
\end{algorithmic}
\end{algorithm}
\begin{algorithm}[htbp]
   \caption{Target choice with the historical strategy}
   \label{alg:targetchoice}
\begin{algorithmic}[1]
   \STATE {\bfseries Input:} a training sample $\bmx_0$, missing pattern dataset $D_{\textrm{miss}}$
   \STATE {\bfseries Output:} conditional information $\bmx_0^\rmo$, imputation targets $\bmx_0^\rmu$
   \STATE Draw a data sample $\tilde{\bmx}_0$ from $D_{\textrm{miss}}$ 
   \STATE Denote the indices of observed values of $\bmx_0$ as $J$
   \STATE Denote the indices of missing values of $\tilde{\bmx}_0$ as $\tilde{J}$
   \STATE Take the intersection of $J$ and $\tilde{J}$, and denote values of $\bmx_0$ for the intersection as $\bmx_0^\rmu$
   \STATE Set the remaining observations of $\bmx_0$ as $\bmx_0^\rmo$
\end{algorithmic}
\end{algorithm}

\subsection{Target choice strategies for self-supervised training}
\label{append:algorithms:knownmask}
We describe the target choice strategies for self-supervised training of \pname, which is discussed in \Secref{sec:model:targetchoice}. We give the algorithm of the random strategy in Algorithm~\ref{alg:randomchoice} and that of the historical strategy in Algorithm~\ref{alg:targetchoice}. 
On the historical strategy, we use the training dataset as missing pattern dataset $D_{\textrm{miss}}$, unless otherwise stated. The mix strategy draws one of the two strategies with a ratio of 1:1 for each training sample. The test pattern strategy just uses the fixed missing pattern in the test dataset to choose imputation targets.

\section{Training and imputation for unconditional diffusion model}
\label{append:uncond}

\subsection{Imputation with unconditional diffusion model}
\label{append:uncondsampling}
We describe the imputation method with the unconditional diffusion model used for the experiments in \Secref{sec:experiments:imp}.
We followed the method described in previous studies~\citep{song2021score}. 
To utilize unconditional diffusion models for imputation, they approximated the conditional reverse process $p_\theta(\bmx_{t-1}^\rmu \mid \bmx_t^\rmu,\bmx_0^\rmo)$ in \eqref{eq:reverse_cond_part} with the unconditional reverse process in \eqref{eq:reverse}. 
Given a test sample $\bmx_0$, 
they set all observed values as conditional observations $\bmx_0^\rmo$ and all missing values as imputation targets $\bmx_0^\rmu$. Then, instead of conditional observations $\bmx_0^\rmo$, they considered noisy conditional observations ${\bmx}_t^\rmo := \sqrt{{\alpha}_t}\bmx_0^\rmo + (1-{\alpha}_t) \rvepsilon$ and exploited ${\bmx}_{t} = [{\bmx}_t^\rmo;{\bmx}_t^\rmu]\in \gX$ for the input to the distribution $ p_\theta({\bmx}_{t-1} \mid {\bmx}_{t})$ in \eqref{eq:reverse}, where $[{\bmx}_t^\rmo;{\bmx}_t^\rmu]$ combines ${\bmx}_t^\rmo$ and ${\bmx}_t^\rmu$ to create a sample in $\gX$. Using this approximation, we can sample ${\bmx}_{t-1}$ from $ p_\theta({\bmx}_{t-1} \mid {\bmx}_{t} =[{\bmx}_t^\rmo;{\bmx}_t^\rmu])$ and obtain ${\bmx}_{t-1}^\rmu$ by extracting target indices from ${\bmx}_{t-1}$. By repeating the sampling procedure from $t=T$ to $t=1$, we can generate imputation targets ${\bmx}_{0}^\rmu$.

\subsection{Training procedure of unconditional diffusion models for time series imputation}
\label{append:uncondtrain}
In \Secref{sec:background:ddpm}, we described the training procedure of the unconditional diffusion model, which expects the training dataset does not contain missing values. However, the training dataset that we use for time series imputation contains missing values. To handle missing values, we slightly modify the training procedure. 
Given a training sample $\bmx_0$ with missing values, 
we treat the missing values like observed values by filling dummy values to the missing indices of $\bmx_0$. We adopt zeros for the dummy values and denote the training sample after filling zeros as $\widehat{\bmx}_0$. 
Since all indices of $\widehat{\bmx}_0$ contain values, we can sample noisy targets $\sqrt{\alpha_t}\widehat{\bmx}_0 + (1-\alpha_t) \rvepsilon$ as with the training procedure in \Secref{sec:background:ddpm}. 
We consider denoising the noisy target for training, but we are only interested in estimating the noises added to the observed indices since the dummy values contain no information about the data distribution. To exclude the missing indices, we introduce an observation mask $\bmM\in \{0,1\}^{K\times L}$, which takes value 1 for observed indices. Then, instead of \eqref{eq:loss}, we use the following loss function for training under the existence of missing values:
\begin{align}
  \min_{\theta} L(\theta) := \min_{\theta} 
  \bb{E}_{\bmx_0 \sim q(\bmx_0), \rvepsilon \sim \gN(\bmzero, \bmI),t}
  ||  (\rvepsilon - \rvepsilon_{\theta}(\sqrt{\alpha_t}\widehat{\bmx}_0 + (1-\alpha_t) \rvepsilon, t)) \odot \bmM ||_2^2.
  \label{eq:loss_uncond}
\end{align}

\begin{figure*}[htbp]
\centering
        \begin{center}
          \includegraphics[width=1.\textwidth]{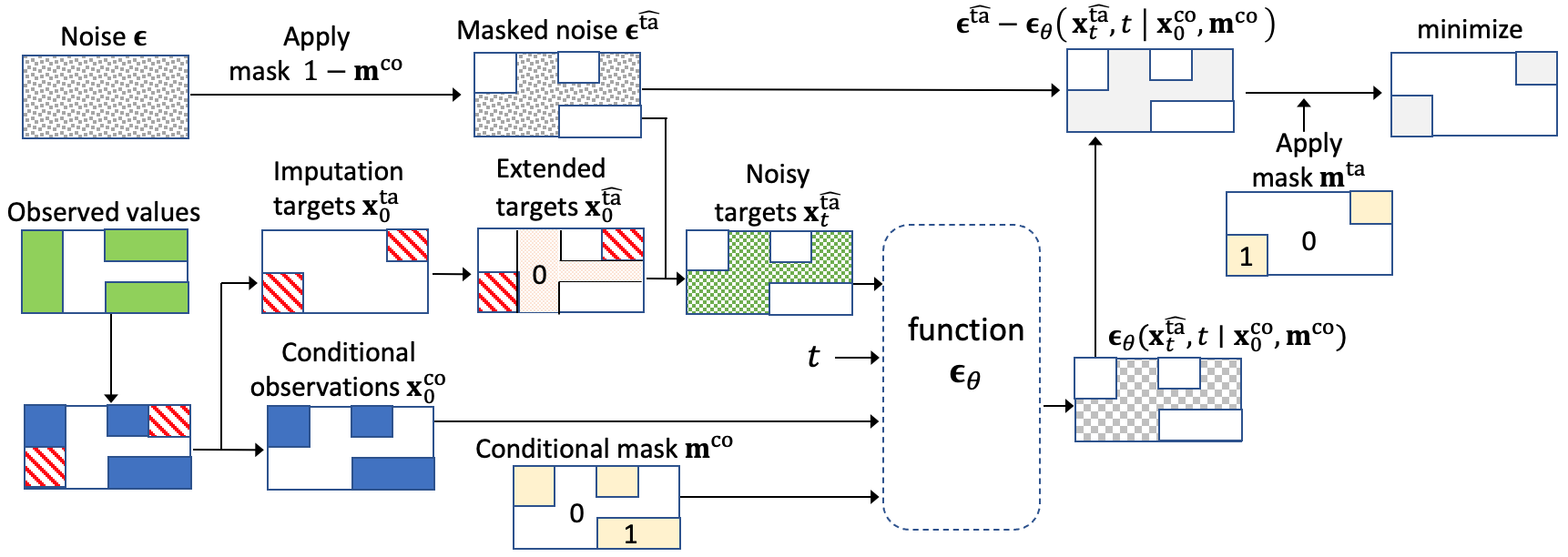}
        \end{center}
\caption{The self-supervised training procedure of \pname{} for implementation of time series imputation. The colored areas in each rectangle represent the existence of values. The green and white areas represent observed and missing values, respectively, and white areas are padded with zeros to fix the shape of the inputs.  Zero padding is also applied to all white areas. As with \Figref{fig:training}, the observed values are separated into red imputation targets $\bmx_0^\rmu$ and blue conditional observations $\bmx_0^\rmo$. For the extended targets $\bmx_0^\hatrmu$, the area of value 0 shows dummy values.}%
    \label{fig:training_extend}
\end{figure*}

\section{\pname{} for implementation of time series imputation}
\label{append:model_extend}
In this section, we discuss the effect of adjusting the function $\rvepsilon_\theta$, described in \Secref{sec:main:time}. 
First, let us consider the effect of the adjustment on sampling. The adjustment does not essentially affect the model at sampling time, because all values of data are either conditional observations or imputation targets as shown in \Tabref{tab:summaryinput} and the model can distinguish the type of each value through the mask $\bmM^\rmo$. 
Since the output shape is adjusted, we need to recover the shape by extracting the indices of the imputation targets from the output, so that we substitute the outputs into \eqref{eq:reverse_cond}.

Next, we focus on the effect of the adjustment on training.
Unlike sampling, the model at training time cannot distinguish imputation targets and missing values since we ignore missing values during training as shown in \Tabref{tab:summaryinput}. 
In order to handle the missing values, we need to modify the inputs to $\rvepsilon_\theta$. 
Here, we use a similar approach to the training procedure with the unconditional model in \Secref{append:uncondtrain}.
Namely, we treat the missing indices like a part of imputation targets. We illustrate the extended training procedure in \Figref{fig:training_extend}. 
First, we set zeros to the missing indices as dummy values. 
We denote the extended imputation targets as $\bmx_0^\hatrmu$. Then, we sample noisy targets $\bmx_t^\hatrmu = \sqrt{{\alpha}_t}\bmx_0^\hatrmu +  (1-{\alpha}_t) \rvepsilon^\hatrmu$, where $\rvepsilon^\hatrmu$ is masked noise and is given by
$\rvepsilon^\hatrmu:=  (1-\bmM^\rmo) \odot \rvepsilon$, as shown in \Figref{fig:training_extend}.
We denoise the noisy targets for training. 
We only estimate the noise for the original imputation targets, since the dummy values contain no information about the data distribution. In other words, we train $\rvepsilon_\theta$ by solving the following optimization problem:
\begin{align}
  \min_{\theta} \gL(\theta) := \min_{\theta} 
  \bb{E}_{\bmx_0 \sim q(\bmx_0), \rvepsilon \sim \gN(\bmzero, \bmI),t}
  || (\rvepsilon - \rvepsilon_\theta (\bmx_t^\hatrmu,t \mid \bmx_0^\rmo, \bmM^\rmo)) \odot \bmM^\rmu ||_2^2
  \label{eq:loss_cond_miss}
\end{align}
where $\bmM^\rmu$ is a mask which corresponds to $\bmx_0^\rmu$ and takes value 1 for the original imputation targets.

\section{Details of architectures and experiment settings}
\label{append:settings}

\begin{figure*}[htbp]
\centering
        \begin{center}
          \includegraphics[width=.95\textwidth]{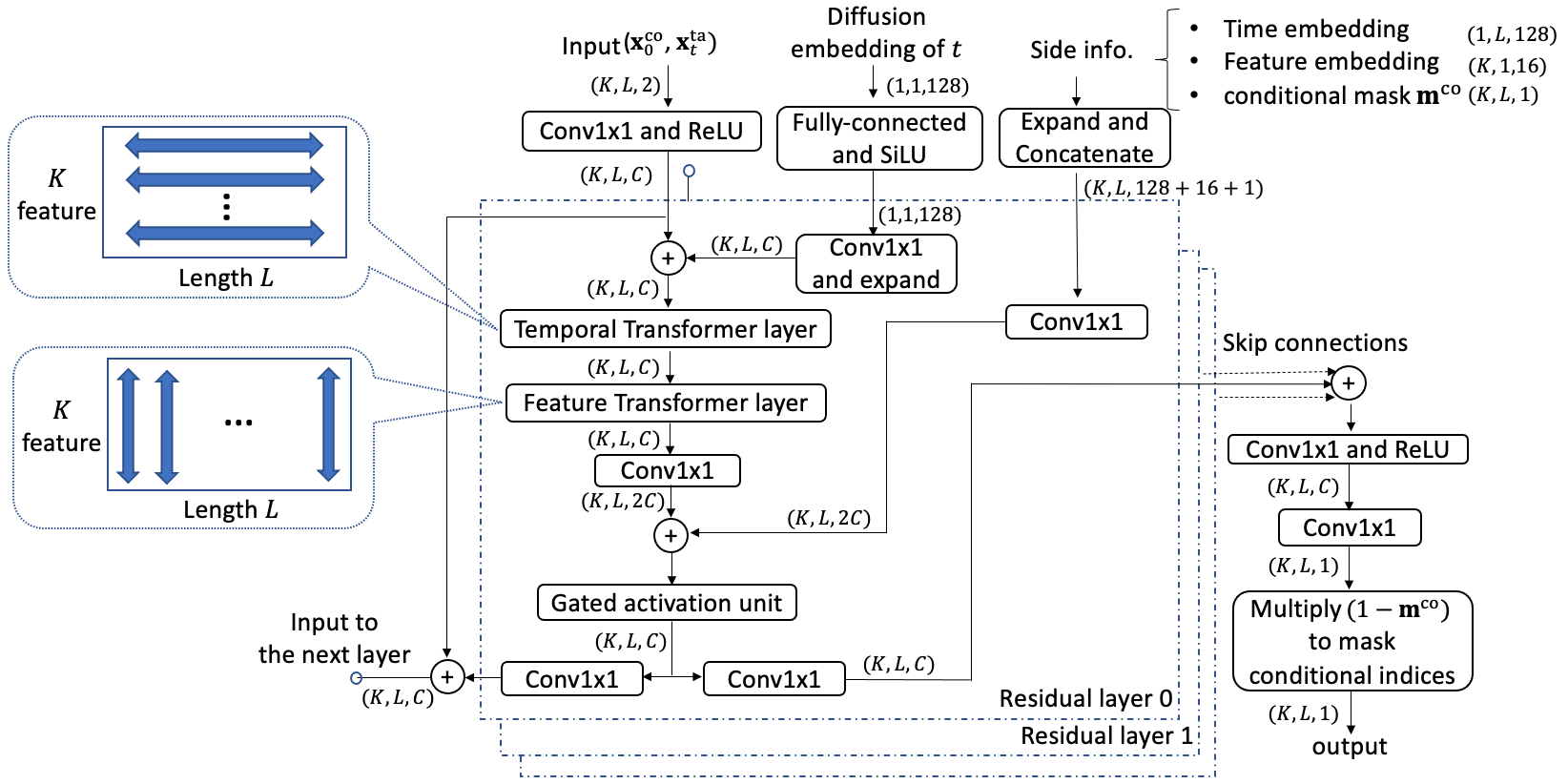}
        \end{center}
    \caption{Architecture of $\rvepsilon_\theta$ in \pname{} for multivariate time series imputation.}
    \label{fig:architecture}
\end{figure*}

\subsection{Details of implementation of \pname }
\label{append:settings:diffusion_model}
We describe the details of architectures and hyperparameters for the conditional diffusion model described in \Secref{sec:main:architecture}. 
First, we provide the whole architecture of \pname{} in \Figref{fig:architecture}. Since the architecture in \Figref{fig:architecture} is based on DiffWave~\citep{kong2020diffwave}, we mainly explain the difference from DiffWave.

On the top of the figure, the models take $\bmx_0^\rmo$ and $\bmx_t^\rmu$ as inputs since $\rvepsilon_\theta$ is the conditional denoising function. 
For the diffusion step $t$, we use the following 128-dimensions embedding following previous works~\citep{vaswani2017attention,kong2020diffwave}:
\begin{align}
t_{embedding}(t) = \left( \sin(10^{0\cdot 4/63} t),\ldots,\sin(10^{63\cdot 4/63} t),\cos(10^{0\cdot 4/63} t),\ldots,\cos(10^{63\cdot 4/63} t)
\right).
\end{align}
Similarly, we utilize time embedding of $\bms=\{s_{1:L}\}$ as a side information. We use 128-dimensions temporal embedding following previous studies~\citep{vaswani2017attention,zuo2020transformer}:
\begin{align}
s_{embedding}(s_l) = \left( \sin(s_l/\tau^{0/64}),\ldots,\sin(s_l/\tau^{63/64}),\cos(s_l/\tau^{0/64}),\ldots,\cos(s_l/\tau^{63/64})
\right)
\end{align}
where $\tau = 10000$. On the top right of the figure, we expand each side information and concatenate all the side information. On the bottom right of the figure, we multiply the output by a mask $(1-\bmM^\rmo)$ in order to mask the indices of the conditional observations of the output.

As for Transformer layers, we used 1-layer TransformerEncoder implemented in PyTorch~\citep{paszke2019pytorch}, which is composed of a multi-head attention layer, fully-connected layers and layer normalization. 
Only for forecasting tasks, we adopted the "linear attention transformer" package~\citep{lineartransformer} to improve computational efficiency, since the forecasting datasets we used contained many features and long sequences.
The package implements an efficient attention mechanism~\citep{shen2021efficient}, and we only used global attention in the package.

\subsection{Details of experiment settings in \Secref{sec:experiments}}
\label{append:settings:experiments}
In this section, we provide the details of the experiment settings in \Secref{sec:experiments}. When we evaluated baseline methods with the original implementation in each section, we used their original hyperparameters and model size. Although we also ran experiments under the same model size as our model, the performance did not improve in more than half of the cases and did not outperform our model in all cases.

\subsubsection{Experiment settings for imputation in \Secref{sec:experiments:imp}}
\label{append:settings:experiment1}
First, we explain additional information for the air quality dataset. The dataset is composed of air quality data in Beijing from 2014/05/01 to 2015/04/30. The dataset contains artificial ground-truth, whose missing patterns are created based on those in the next month. 

Next, we describe data splits. For the healthcare dataset, we randomly divided the dataset into five parts and used one of them as test data for each run. We also randomly split the remaining data into train and validation data with a ratio of 7:1. 
For the air quality dataset, following~\citep{yi2016st}, we used the 3rd, 6th, 9th and 12th months as test data. To avoid evaluating imputation for each missing value multiple times, we separated the test data of each month every 36 consecutive time steps without overlap. When the length of a monthly data was not divisible by 36, we allowed the last sequence to overlap with the previous one and did not aggregate the result for the overlapping parts. For each run, we selected a month as validation data and used the rest as training data. 
We note that we excluded 
the 4th, 7th, 10th, and 1st months from missing pattern dataset for the historical strategy, because these months were used for creating missing patterns of the artificial ground-truth. 

On the healthcare dataset, due to the different scales of features, we evaluate the performance on normalized data following previous studies~\citep{cao2018brits}. For training of all tasks, we normalize each feature to have zero mean and unit variance.

As for hyperparameters, we set the batch size as 16 and the number of epochs as 200. We used Adam optimizer with learning rate $0.001$ that is decayed to $0.0001$ and $0.00001$ at 75\% and 90\% of the total epochs, respectively. As for the model, we set the number of residual layers as 4, residual channels as 64, and attention heads as 8. We followed DiffWave\citep{kong2020diffwave} for the number of channels and decided the number of layers based on the validation loss and the parameter size. The number of the parameter  in the model is about 415,000.

We also provide hyperparameters for the diffusion model as follows. We set the number of the diffusion step $T=50$, the minimum noise level $\beta_1=0.0001$, and the maximum noise level $\beta_{T}=0.5$. 
Since recent studies\citep{song2020DDIM,nichol2021improved} reported that gentle decay of $\alpha_t$ could improve the sample quality, we adopted the following quadratic schedule for other noise levels:
\begin{align}
\label{eq:quadschedule}
\beta_t
=\left( \frac{T-t}{T-1}\sqrt{\beta_1} +
\frac{t-1}{T-1}\sqrt{\beta_{T}} \right)^2. 
\end{align}

With regard to the baselines for probabilistic imputation,
we used their original implementations for GP-VAE and V-RIN. For Multitask GP, we utilized GPyTorch~\citep{gardner2018gpytorch} for the implementation. We used RBF kernel for the covariance between timepoints and low-rank IndexKernel with $rank = 10$ for that between features.

Finally, we describe the baselines for deterministic imputation, which were used for comparison. 
1) BRITS~\citep{cao2018brits}: the method utilizes a bi-directional recurrent neural network to handle multiple correlated missing values. 
2) V-RIN~\citep{mulyadi2021uncertainty}: the method utilizes the uncertainty learned with VAE to improve recurrent imputation. 
3) GLIMA~\citep{suo2020glima}: the method combines recurrent imputations with an attention mechanism to capture cross-time and cross-feature dependencies and shows the state-of-the-art performance. 
4) RDIS~\citep{choi2020rdis}: the method applies random drops to given training data for self-training.
We used the original implementation for BRITS and V-RIN. For RDIS, we set the number of models as 8, hidden units as 108, drop rate as 30\%, threshold as 0.1, update epoch as 200, and total epochs as 1000.

\subsubsection{Experiment settings for interpolation in \Secref{sec:experiments:interp}}
\label{append:settings:experiment2}
First, we explain how we process the dataset. We processed the healthcare dataset as irregularly sampled time series. Following previous studies~\citep{shukla2021multi,rubanova2019latentODE}, we rounded observation times to the nearest minute. Then, there are $48\times 60$ possible measurement times per time series, and the lengths of time series samples can be different each other.

We used almost the same experiment settings as those for imputation in \Secref{append:settings:experiment1}. Since the length of each irregularly sampled time series is different, we applied zero padding to each time series in order to fix the length for each batch. The padding does not affect the result since the attention mechanisms in the implementation of \pname{} can deal with the padding by using padding masks. 

We describe the baselines which were used for comparison. We used the original implementation.
1) Latent ODE~\citep{rubanova2019latentODE}: the method consists of an ODE-RNN model as the encoder and a neural ODE model as the decoder. 
2) mTANs~\citep{shukla2021multi}: the method utilized an attention mechanism and showed state-of-the-art results for the interpolation of irregularly sampled time series.

\begin{table*}[htb]
\caption{Description of datasets for time series forecasting.}
\label{tab:data_forecasting}
\begin{center}
\setlength{\tabcolsep}{4pt}
\begin{tabular}{lcccccc}
\toprule & feature $K$ & 
\begin{tabular}{c}
total \\ time step \end{tabular} & 
\begin{tabular}{c} history \\ steps $L_1$  \end{tabular} &
\begin{tabular}{c} prediction \\ steps $L_2$  \end{tabular} &
\begin{tabular}{c} test \\ sample  \end{tabular} & epochs \\ \midrule 
 solar & 137 & 10392 & 168 & 24 & 7 & 50 \\
 electricity & 370 & 5833 & 168 & 24 & 7 & 100 \\
 traffic & 963 &  7009 & 168 & 24 & 7 & 200 \\
 taxi & 1214 &  1488 & 48 & 24 & 56 & 300 \\
 wiki & 2000 &  792 & 90 & 30 & 5 & 300 \\
\bottomrule
\end{tabular}
\end{center}
\end{table*}

\subsubsection{Datasets and Experiment settings for forecasting in \Secref{sec:experiments:forecast}}
\label{append:settings:experiment3}
First we describe the datasets we used. 
We used five open datasets that are commonly used for evaluating probabilistic time series forecasting. The datasets were preprocessed in Salinas et al.~\citep{salinas2019high} and provided in GluonTS\footnote{\url{https://github.com/awslabs/gluon-ts}}\citep{gluonts_jmlr}:
\begin{itemize}
    \item solar~\citep{lai2018solar}: hourly solar power production records of 137 stations in Alabama State.
    \item electricity\footnote{\url{https://archive.ics.uci.edu/ml/datasets/ElectricityLoadDiagrams20112014}}: hourly electricity consumption of 370 customers.
    \item traffic\footnote{\url{https://archive.ics.uci.edu/ml/datasets/PEMS-SF}}: hourly occupancy rate of 963 San Fancisco freeway car lanes.
    \item taxi\footnote{\url{https://www1.nyc.gov/ site/tlc/about/tlc-trip-record-data}}: half hourly traffic time series of New York taxi rides taken at 1214 locations in the months of January 2015 for training and January 2016 for test.
    \item wiki: daily page views of 2000 Wikipedia pages.
\end{itemize}
We summarize the characteristics of each dataset in \Tabref{tab:data_forecasting}. 
The task for these datasets is to predict the future $L_2$ steps by exploiting the latest $L_1$ steps where $L_1$ and $L_2$ depend on datasets as shown in \Tabref{tab:data_forecasting}. We set $L_1$ and $L_2$ referring to previous studies~\citep{nguyen2021temporal}. 
For training, we randomly selected $L_1+L_2$ consecutive time steps as one time series and set the last $L_2$ steps as imputation targets. 
We followed the train/test split in previous studies. We used the last five samples of training data as validation data. 

As for experiment settings, since we basically followed the setting for time series imputation in \Secref{append:settings:experiment1}, we only describe the difference from it. We ran each experiment three times with different random seeds. We set batch size as 8 because of longer sequence length, and utilized an efficient Transformer as mentioned in \Secref{append:settings:diffusion_model}.

Since the number of features $K$ is large, we adopted subset sampling of features for training. For each time series in a training batch, we randomly chose a subset of features and only used the subset for the batch. The attention mechanism allows the model to take varying length inputs. We set the subset size as $64$. Due to the subset sampling, we need large epochs when the number of features $K$ is large. Therefore, we set training epochs based on the number of features and the validation loss. We provide the epochs in \Tabref{tab:data_forecasting}.

Finally, we describe the baselines which were used for comparison. 
1) GP-copula~\citep{salinas2019high}: the method combines a RNN-based model with a Gaussian copula process to model time-varying correlations. 
2) Transformer MAF~\citep{rasul2020multi}: the method uses Transformer to learn temporal dynamics and a conditioned normalizing flow to capture feature dependencies. 
3) TLAE~\citep{nguyen2021temporal}: the method combines a RNN-based model with autoencoders to learn latent temporal patterns. 
4) TimeGrad~\citep{rasul2021autoregressive}: the method has shown the state-of-the-art results for probabilistic forecasting by combining a RNN-based model with diffusion models.

\subsection{Computations of CRPS}
\label{append:CRPS}
We describe the definition and computation of the CRPS metric.

The continuous ranked probability score (CRPS)~\citep{matheson1976CRPS} measures the compatibility of an estimated probability distribution $F$ with an observation $x$, and can be defined as the integral of the quantile loss $\Lambda_\alpha(q,z):= (\alpha-\1_{z<q})(z-q)$ for all quantile levels $\alpha \in [0,1]$:
\begin{align}
    \textrm{CRPS}(F^{-1},x)=\int_0^1 2 \Lambda_\alpha (F^{-1}(\alpha),x)\mathrm{d}\alpha
    \label{eq:CRPS}
\end{align}
where $\1$ is the indicator function. 
We generated 100 samples to approximate the distribution $F$ over each missing value. 
We computed quantile losses for discretized quantile levels with $0.05$ ticks. Namely, we approximated CRPS with
\begin{align}
    \textrm{CRPS}(F^{-1},x)\simeq \sum_{i=1}^{19} 2 \Lambda_{i*0.05} (F^{-1}(i*0.05),x) / 19.
    \label{eq:CRPSsum}
\end{align}
Then, we evaluated the following normalized average of CRPS for all features and time steps:
\begin{align}
    \frac{\sum_{k,l}\textrm{CRPS}(F^{-1}_{k,l},x_{k,l})}
    {\sum_{k,l}|x_{k,l}|}
\end{align}
where $k$ and $l$ indicates features and time steps of imputation targets, respectively.

For probabilistic forecasting, we evaluated CRPS-sum. CRPS-sum is CRPS for the distribution $F$ of the sum of all $K$ features and is computed by the following equation:
\begin{align}
    \frac{\sum_{l}\textrm{CRPS}(F^{-1},\sum_{k} x_{k,l})}
    {\sum_{k,l}|x_{k,l}|}
\end{align}
where $\sum_{k} x_{k,l}$ is the sum of forecasting targets for all features at time point $l$.

\section{Additional results and experiments}
\begin{table*}[htbp]
\caption{Comparing the two dimension attention mechanism of various architectures. For ablations, we report the mean and the standard error for three trials.
}
\begin{center}
\begin{tabular}{lcccc} 
\toprule
& \multicolumn{2}{c}{healthcare  ($10 \%$ missing)} & \multicolumn{2}{c}{air quality} \\
& MAE & CRPS & MAE & CRPS \\
\midrule
 no-temporal & $0.439(0.004)$ & $0.475(0.001)$ & $26.63(0.23)$ & $0.292(0.002)$ \\
 no-feature & $0.352(0.001)$ & $0.386(0.002)$ & $14.44(0.11)$ & $0.162(0.001)$ \\
 flatten & $0.383(0.002)$ & $0.418(0.002)$ & $12.26(0.09)$ & $0.139(0.001)$ \\
 Bi-RNN & $0.272(0.001)$ & $0.301(0.001)$ & $12.56(0.26)$ & $0.142(0.003)$ \\
 dilated conv & $0.279(0.002)$ & $0.305(0.002)$ & $11.67(0.11)$ & $0.130(0.001)$ \\
 2D attention (proposed) & ${\bf 0.217(0.001) }$ & ${\bf 0.238(0.001) }$ & ${\bf 9.60(0.04) }$ & ${\bf 0.108(0.001) }$ \\
\bottomrule
\end{tabular}
\end{center}
\label{tab:ablation}
\end{table*}

\subsection{Effectiveness of two dimensional attention mechanism}
\label{append:ablation_attention}
In this paper, we utilized a two dimensional attention mechanism to learn temporal and feature dependencies. To show the effectiveness of the attention mechanism, we demonstrate an ablation study. We replace the attention mechanism with the following architecture baselines and compare the performance:
\begin{itemize}
    \item no temporal: remove temporal attention layers
    \item no feature: remove feature attention layers
    \item flatten: flatten 2D tensor ($K$ features x $L$ length) to 1D, and input the 1D vector to transformer layers
    \item Bi-RNN: replace the attention mechanism with Bi-directional RNN which is a popular architecture for multivariate time series imputation
    \item dilated conv: replace temporal and feature attention layers with 1D dilated convolution layers, respectively. The dilated convolution was used in previous studies for diffusion models\cite{kong2020diffwave,rasul2021autoregressive}
\end{itemize}
We set hyperparameters of each architecture so that the number of parameters is almost the same as our attention mechanism. 
We show the result in \Tabref{tab:ablation}. Our attention mechanism outperforms all of the other architectures. The comparison with “no temporal” and “no feature” shows that both temporal and feature correlations are important for accurate imputation. The comparison with “flatten”, “Bi-RNN”, and “dilated conv” shows that our attention mechanism is effective to learn temporal and feature dependency compared with existing methods. In summary, the result of the ablation indicates the proposed attention mechanism plays a key role in improving the imputation performance by a large margin.

\begin{table*}[htbp]
\caption{Comparison of the negative log likelihood (NLL) and CRPS for various schedules. We report the mean for three trials.
}
\begin{center}
\begin{tabular}{llcccc} 
\toprule
&& \multicolumn{2}{c} { healthcare  ($10 \%$ missing) } & \multicolumn{2}{c} { air quality } \\
method & schedule & NLL & CRPS & NLL & CRPS \\
\midrule
GP-VAE & $-$ & $<1.22$ & $0.574$ & $<1.09$ & $0.397$ \\
\hline proposed & quad. (in paper) & $<1.63$ & ${\bf 0.238}$ & $<0.97$ & ${\bf 0.108}$ \\
\hline proposed & linear & $<29.70$ & $0.240$ & $<18.55$ & $0.110$ \\
\hline proposed & quad. (large min. noise) & ${\bf<0.07}$ & $0.239$ & ${\bf<-0.70}$ & $0.109$
\\
\bottomrule
\end{tabular}
\end{center}
\label{tab:NLL}
\end{table*}

\subsection{Comparison of negative log likelihood for probabilistic imputation}
The negative log likelihood (NLL) is a popular metric for evaluating probabilistic methods and ELBO is often utilized to estimate NLL. A reason why we mainly focused on other metrics is that ELBO is sometimes far from NLL and uncorrelated with the quality of generated samples. Specifically, in the proposed method, the choice of the noise schedule highly affects the ELBO while it has little effect on the sample quality.

To demonstrate this, we performed an experiment. We chose the following three noise schedules for CSDI and calculated NLL and CRPS for each schedule.
\begin{itemize}
    \item quadratic (used in the paper): quadratic spaced schedule between $\beta_{\min} = 0.0001$ and $\beta_{\max} = 0.5$
    \item linear: linear spaced schedule with the same  $\beta_{\min}$ and $\beta_{\max}$ as those in the paper
    \item quadratic (large minimum noise): quadratic schedule with large minimum noise level  $\beta_{\min} = 0.001$, which makes the model ignore small noise
\end{itemize}
We also calculated the metrics for GP-VAE. The result is shown in~\Tabref{tab:NLL}. While CRPS by the proposed method is almost independent from the choice of schedules, NLL significantly depends on the schedule. This phenomenon happens because time series data is generally noisy and it is difficult to denoise small noise during imputation. Estimated scores by the model could be inaccurate when inputs to the model (i.e. imputation targets) only contain small noise. These inaccurate scores could make the estimated ELBO loose, whereas small noise does not affect the sample quality. When the minimum noise level $\beta_{\min}$ is large, since the model does not denoise small noise in sampling steps, ELBO by the proposed method is tightly estimated and smaller than that by GP-VAE. Therefore, ELBO is not suitable for evaluating the sample quality and we adopted other metrics such as CRPS and MAE.

\subsection{Experimental results for other metrics in \Secref{sec:experiments}}
\label{append:other_metrics}
We show the experimental results in \Secref{sec:experiments} for different metrics in \Tabref{tab:result_det_rmse} to \ref{tab:result_forecast_MAE}.
\Tabref{tab:result_det_rmse} evaluates RMSE for deterministic imputation methods. We added SSGAN~\cite{miao2021SSGAN} as an additional baseline, which has shown the state-of-the-art performance for RMSE in the healthcare dataset. We can confirm that \pname{} outperforms all baselines for RMSE. The advantage of \pname{} is particularly large when the missing ratio is low. This result is consistent with that in \Secref{sec:experiments:imp}.

\Tabref{tab:result_interp_rmse} evaluates MAE and RMSE for interpolation methods. The result is consistent with \Tabref{tab:result_interp}. 
\Tabref{tab:result_forecast_CRPS} and~\ref{tab:result_forecast_MAE} report CRPS and MSE for probabilistic forecasting methods, respectively. We exclude TimeGrad~\citep{rasul2021autoregressive} from the baselines, as they did not report these metrics. We can see that \pname{} is competitive with baselines for these metrics as with CRPS-sum.

\begin{table*}[htbp]
\caption{Comparing deterministic imputation methods with \pname{} for RMSE. The results correspond to \Tabref{tab:result_det}. We report the mean and the standard error for five trials. The asterisk means the values are cited from the original paper.
}
\begin{center}
\begin{tabular}{lccccc} 
\toprule
& \multicolumn{3}{c} { healthcare } & air quality \\
\cmidrule{2-4} & $10 \%$ missing & $50 \%$ missing & $90 \%$ missing & \\
\midrule V-RIN~\citep{mulyadi2021uncertainty} & $0.628(0.025)$ & $0.693(0.022)$ & $0.928(0.013)$ & $40.11(1.14)$ \\
 BRITS~\citep{cao2018brits} & $0.619(0.022)$ & $0.693(0.023)$ & $0.836(0.015)$ & $24.47(0.73)$ \\
 RDIS~\citep{choi2020rdis} & $0.633(0.021)$ & $0.741(0.018)$ &	$0.934(0.013)$& $37.49(0.28)$ \\
 SSGAN~\citep{miao2021SSGAN} (*) & $0.598$ & $0.762$ & $0.818$ & $-$ \\
 unconditional & $0.621(0.020)$ & $0.734(0.024)$ & $0.940(0.018)$ & $22.58(0.23)$ \\
{\bf \pname} (proposed) & ${\bf 0.498(0.020)}$ & ${\bf 0.614(0.017)}$ & ${\bf 0.803(0.012)}$ & ${\bf 19.30(0.13)}$ \\
\bottomrule
\end{tabular}
\end{center}
\label{tab:result_det_rmse}
\end{table*}

\begin{table*}[htbp]
\caption{Comparing the state-of-the-art interpolation method with \pname{} for MAE and RMSE. The results correspond to \Tabref{tab:result_interp}. We report the mean and the standard error for five trials.}
\begin{center}
\begin{tabular}{llccc}
\toprule
&& $10 \%$ missing& $50 \%$ missing & $90 \%$ missing \\ \midrule
\multirow{3}{*}{MAE}& 
 Latent ODE~\citep{rubanova2019latentODE} & $0.522(0.002)$ & $0.506(0.003)$ & $0.578(0.009)$ \\
&mTANs~\citep{shukla2021multi} & $0.389(0.003)$ &$0.422(0.003)$ & $0.533(0.005)$ \\ 
& {\bf \pname} (proposed) & ${\bf 0.362(0.001)}$ &  ${\bf 0.394(0.002)}$ & ${\bf 0.518(0.003)}$\\
\midrule
\multirow{3}{*}{RMSE}& 
Latent ODE~\citep{rubanova2019latentODE} & $0.799(0.012)$ & $0.783(0.012)$ & $0.865(0.017)$ \\
&mTANs~\citep{shukla2021multi} & $0.749(0.037)$ &$0.721(0.014)$ & ${\bf 0.836(0.018)}$ \\ 
& {\bf \pname} (proposed) & ${\bf 0.722(0.043)}$ &  ${\bf 0.700(0.013)}$ & ${0.839(0.009)}$\\
\bottomrule
\end{tabular}
\end{center}
\label{tab:result_interp_rmse}
\end{table*}

\begin{table*}[htbp]
\caption{Comparing probabilistic forecasting methods with \pname{} for CRPS. The results correspond to \Tabref{tab:result_forecast}. We report the mean and the standard error for three trials. The results for baseline methods are cited from their paper. 'TransMAF' is the abbreviation for 'Transformer MAF'.}
\label{tab:result_forecast_CRPS}

\begin{center}
\setlength{\tabcolsep}{2.5pt}
\begin{tabular}{lccccc}
 \toprule & solar & electricity & traffic & taxi & wiki \\ \midrule
GP-copula~\citep{salinas2019high} & $0.371(0.022)$ & $0.056(0.002)$ & $0.133(0.001)$ & $0.360(0.201)$ & $0.236(0.000)$ \\
 TransMAF~\citep{rasul2020multi} & $0.368(0.001)$ & $0.052(0.000)$ & $0.134(0.001)$ & $0.377(0.002)$ & $0.274(0.007)$ \\
 TLAE~\citep{nguyen2021temporal} & $\mathbf{0.335(0.025)}$ & $0.058(0.002)$ & $0.097(0.001)$ & $0.369(0.006)$ & $0.298(0.001)$ \\
  {\bf \pname} (proposed) & $0.338(0.012)$ & $\mathbf{0.041(0.000)}$ & $\mathbf{0.073(0.000)}$ & $\mathbf{0.271(0.001)}$ & $\mathbf{0.207(0.002)}$ \\ 
\bottomrule
\end{tabular}
\end{center}
\end{table*}

\begin{table*}[htbp]
\caption{Comparing probabilistic forecasting methods with \pname{} for MSE. The results correspond to \Tabref{tab:result_forecast}. We report the mean and the standard error for three trials.  The results for baseline methods are cited from their paper. 'TransMAF' is the abbreviation for 'Transformer MAF'. 'TransMAF' did not report the standard error.}
\label{tab:result_forecast_MAE}

\begin{center}
\setlength{\tabcolsep}{3.pt}
\begin{tabular}{lccccc}
\toprule & solar & electricity & traffic & taxi & wiki \\ \midrule
GP-copula~\citep{salinas2019high} & 9.8e2(5.2e1) & 2.4e5(5.5e4) & 6.9e-4(2.2e-5) & 3.1e1(1.4e0) & 4.0e7(1.6e9) \\
 TransMAF~\citep{rasul2020multi} & 9.3e2 & 2.0e5 & 5.0e-4 & 4.5e1 & \bf{3.1e7} \\
 TLAE~\citep{nguyen2021temporal} &\bf{6.8e2(7.5e1)} & 2.0e5(9.2e4) & 4.0e-4(2.9e-6) & 2.6e1(8.1e-1) & 3.8e7(4.2e4) \\
 {\bf \pname} (proposed) &9.0e2(6.1e1) & \bf{1.1e5(2.8e3)} & \bf{3.5e-4(7.0e-7)}& \bf{1.7e1(6.8e-2)} & 3.5e7(4.4e4) \\
\bottomrule
\end{tabular}
\end{center}
\end{table*}

\begin{figure}[htbp]
\centering

        \begin{center}
          \includegraphics[width=1.0\textwidth]{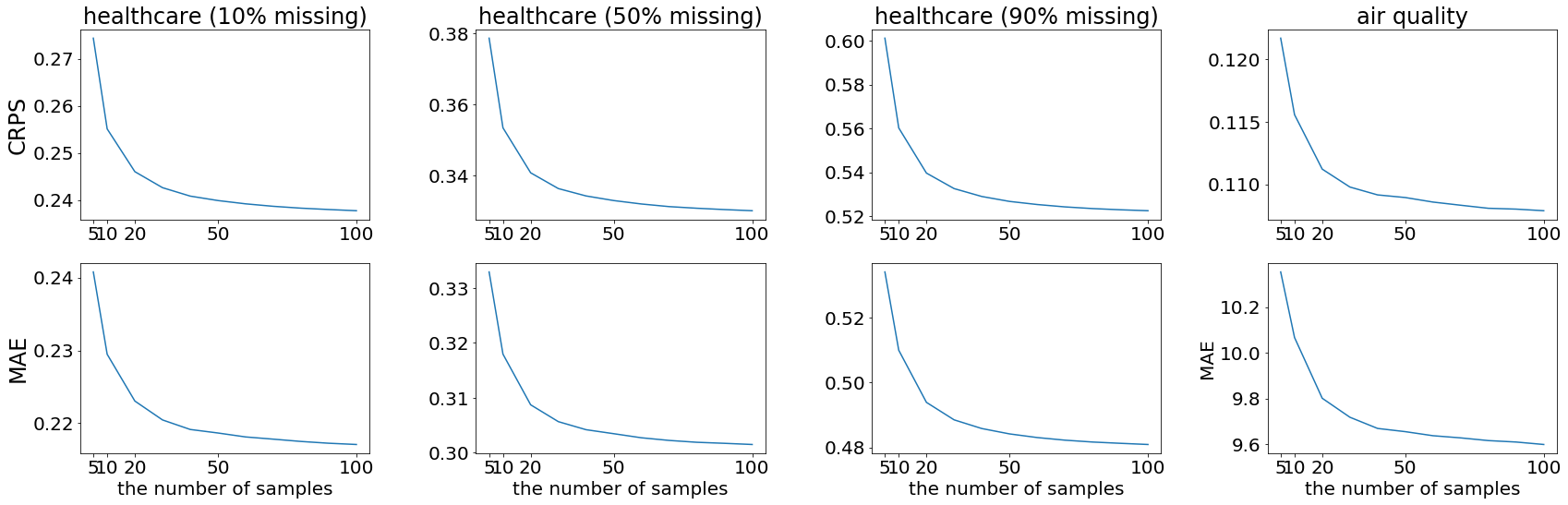}
        \end{center}
    \caption{The effect of the number of generated samples. The first row shows the effect on probabilistic imputation in \Tabref{tab:result_prob} and the second row shows the effect on deterministic imputation in \Tabref{tab:result_det}. } %
    \label{fig:exp_num_samples}
\end{figure}

\subsection{Effect of the number of generated samples}
\label{append:exp_num_samples}
For the experiments in \Secref{sec:experiments}, we generated 100 samples to estimate the distribution of imputation.
We demonstrate the relationship between the number of samples and the performance in \Figref{fig:exp_num_samples}. We can see that five or ten samples are enough to estimate good distributions and outperform the baselines. Increasing the number of samples further improves the performance, and the improvement becomes marginal over 50 samples.

\begin{table*}[htbp]
\caption{The effect of the target choice strategy for the air quality dataset. We report the mean and the standard error for five trials. }
\begin{center}
\begin{tabular}{lcc} 
\toprule
&  { CRPS } & MAE \\
\midrule 
 random & ${\bf 0.108(0.001)}$ & ${\bf 9.58(0.08)}$ \\
 historical &$0.113(0.001)$  & $10.12(0.05)$ \\
 mix & ${\bf 0.108(0.001)}$ & $9.60(0.04)$ \\
\bottomrule
\end{tabular}	
\end{center}
\label{tab:result_choicestrategy}
\end{table*}

\subsection{Effect of target choice strategy}
\label{append:effect_targetchoice}
In the experiment for the air quality dataset in \Secref{sec:experiments:imp}, we adopted the mix strategy for the target choice. Here, we provide the result for other strategies and show the effect of the target choice strategy on imputation quality. 
In \Tabref{tab:result_choicestrategy}, the performances of the mix strategy and the random strategy are almost the same, and the performance of the historical strategy is slightly worse than that of the other strategies. This means that the historical strategy is not effective for the air quality dataset even though the dataset contains structured missing patterns. This is due to the difference of missing patterns between training dataset and test dataset. Note that all strategies outperform the baselines in \Tabref{tab:result_prob} and \Tabref{tab:result_det}.

\section{Additional examples of probabilistic imputation}
\label{append:examples}
\begin{figure}[htbp]
\centering

        \begin{center}
          \includegraphics[width=1.0\textwidth]{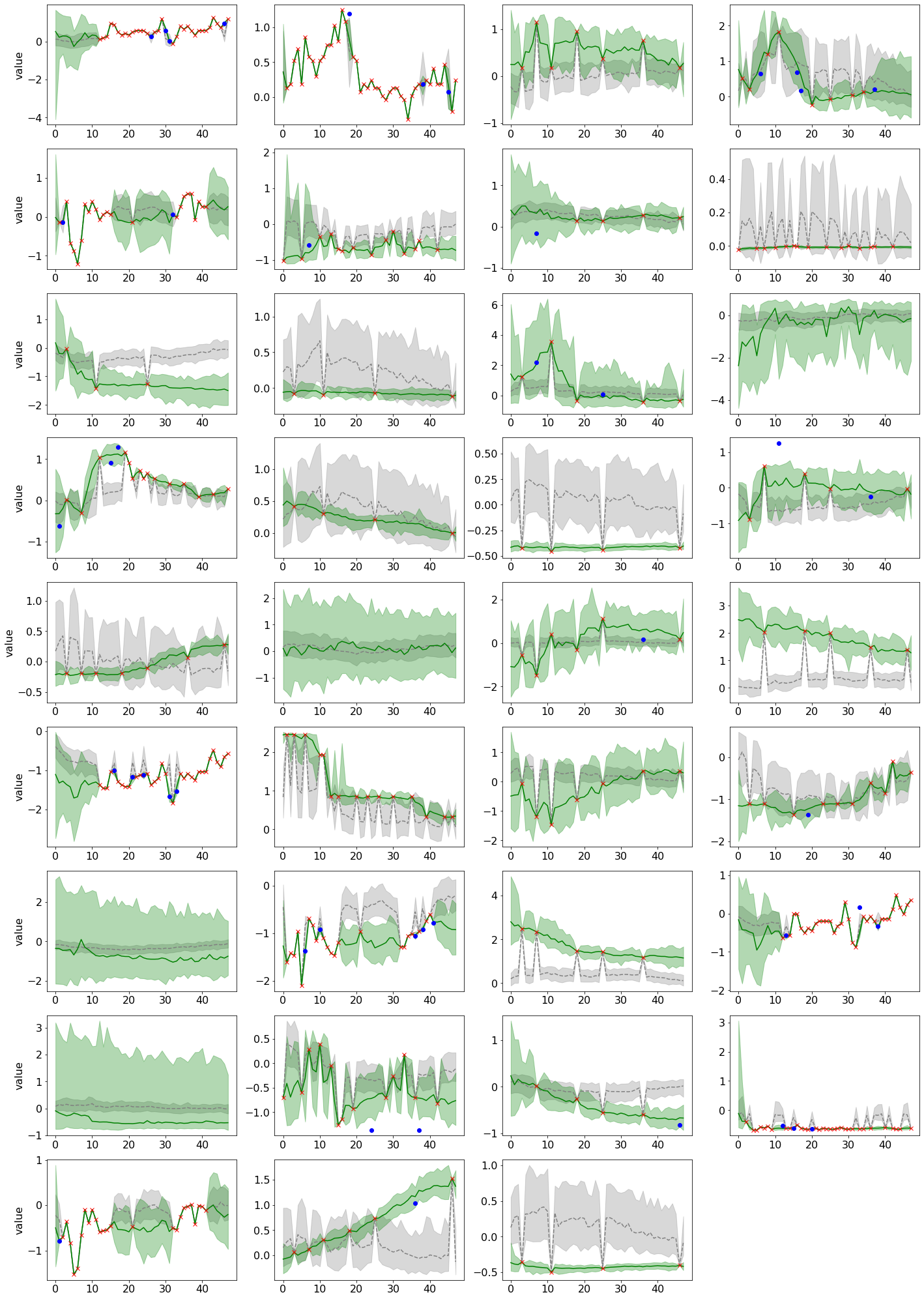}
        \end{center}
    \caption{Comparison of imputation between GP-VAE and \pname{} for the healthcare dataset (10\% missing). The result is for a time series sample with all 35 features. The red crosses show observed values and the blue circles show ground-truth imputation targets. Green and gray colors correspond to \pname{} and GP-VAE, respectively. For each method, median values of imputations are shown as the line and 5\% and 95\% quantiles are shown as the shade.} %
    \label{fig:example_10percent}
\end{figure}

\begin{figure}[htbp]
\centering

        \begin{center}
          \includegraphics[width=1.0\textwidth]{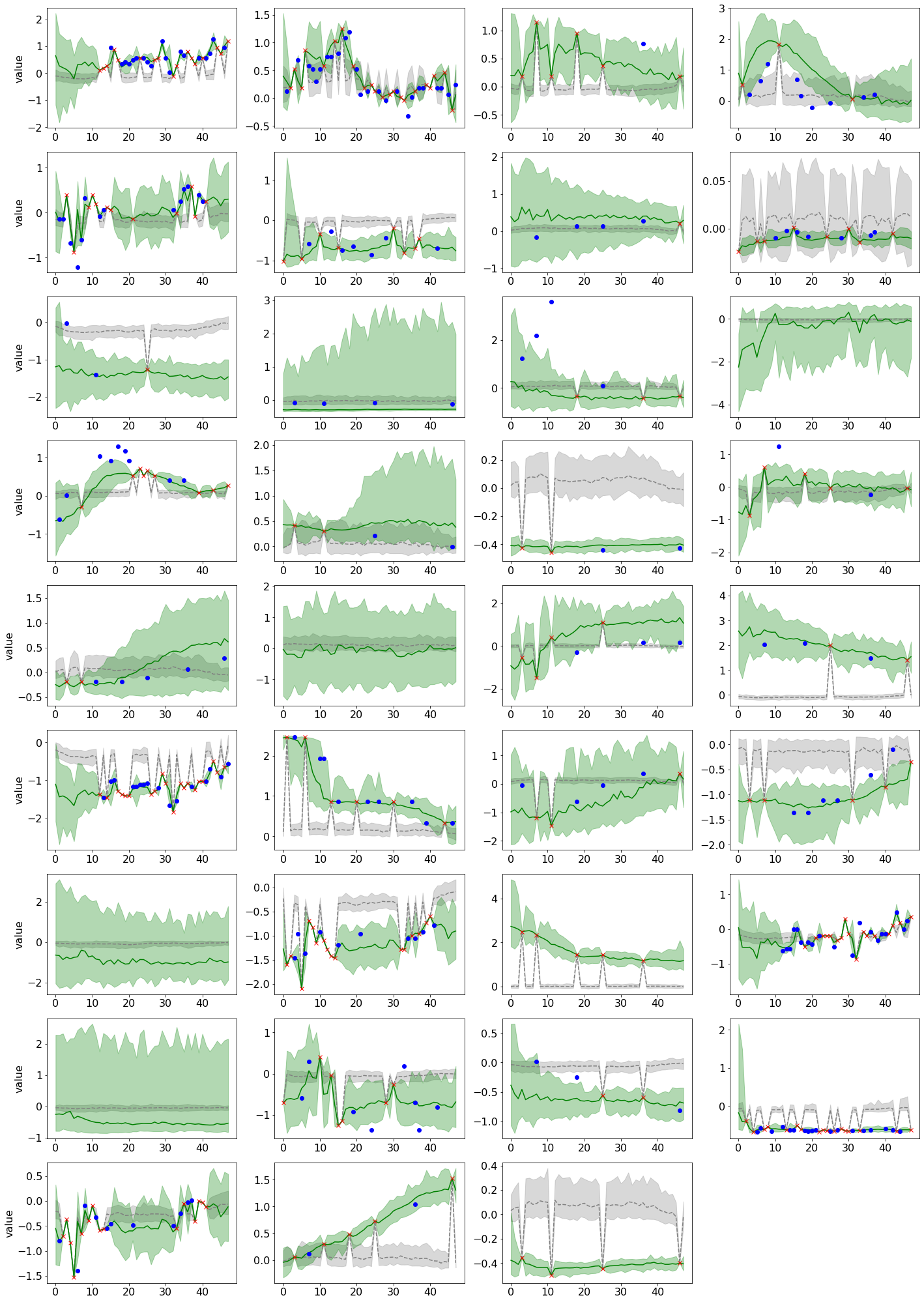}
        \end{center}
    \caption{Comparison of imputation between GP-VAE and \pname{} for the healthcare dataset (50\% missing). The result is for a time series sample with all 35 features. The red crosses show observed values and the blue circles show ground-truth imputation targets. Green and gray colors correspond to \pname and GP-VAE, respectively. For each method, median values of imputations are shown as the line and 5\% and 95\% quantiles are shown as the shade.} %
    \label{fig:example_50percent}
\end{figure}

\begin{figure}[htbp]
\centering

        \begin{center}
          \includegraphics[width=1.0\textwidth]{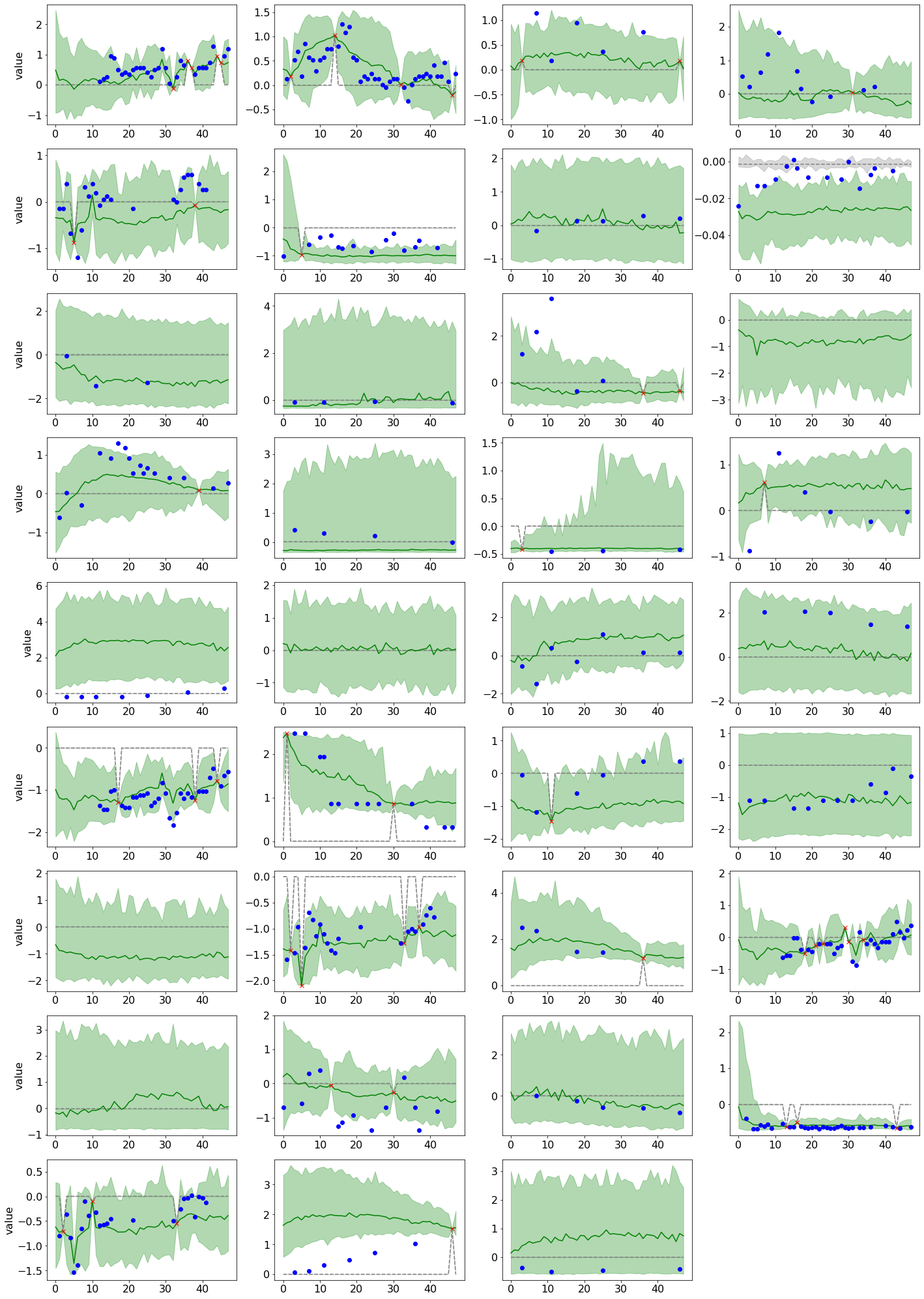}
        \end{center}
    \caption{Comparison of imputation between GP-VAE and \pname{} for the healthcare dataset (90\% missing). The result is for a time series sample with all 35 features. The red crosses show observed values and the blue circles show ground-truth imputation targets. Green and gray colors correspond to \pname{} and GP-VAE, respectively. For each method, median values of imputations are shown as the line and 5\% and 95\% quantiles are shown as the shade.} %
    \label{fig:example_90percent}
\end{figure}

\begin{figure}[htbp]
\centering

        \begin{center}
          \includegraphics[width=1.0\textwidth]{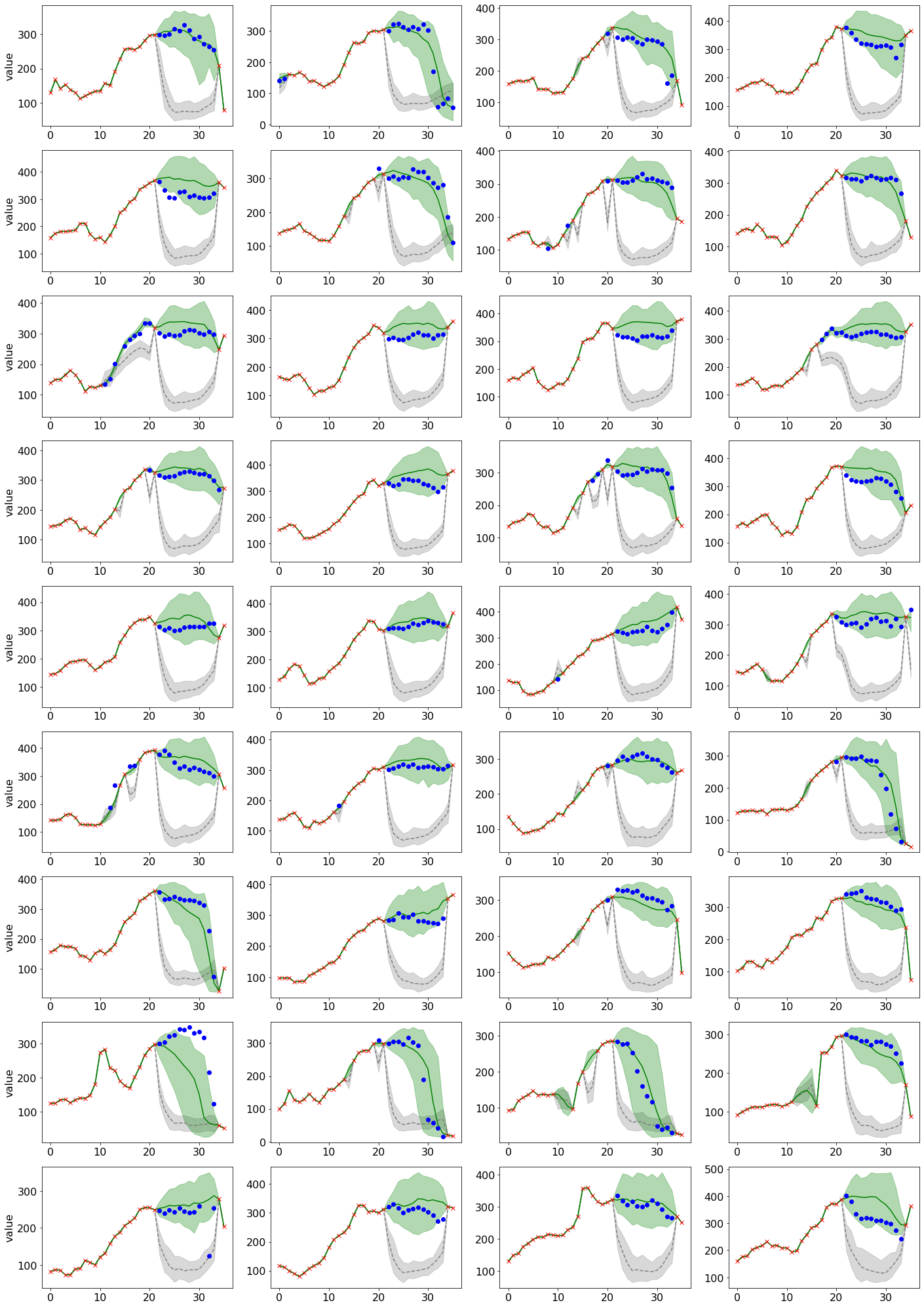}
        \end{center}
    \caption{Comparison of imputation between GP-VAE and \pname{} for the air quality dataset. The result is for a time series sample with all 36 features. 
    The red crosses show observed values and the blue circles show ground-truth imputation targets. Green and gray colors correspond to \pname{} and GP-VAE, respectively. For each method, median values of imputations are shown as the line and 5\% and 95\% quantiles are shown as the shade. } %
    \label{fig:example_airquality}
\end{figure}

\begin{figure}[htbp]
\centering

        \begin{center}
          \includegraphics[width=1.0\textwidth]{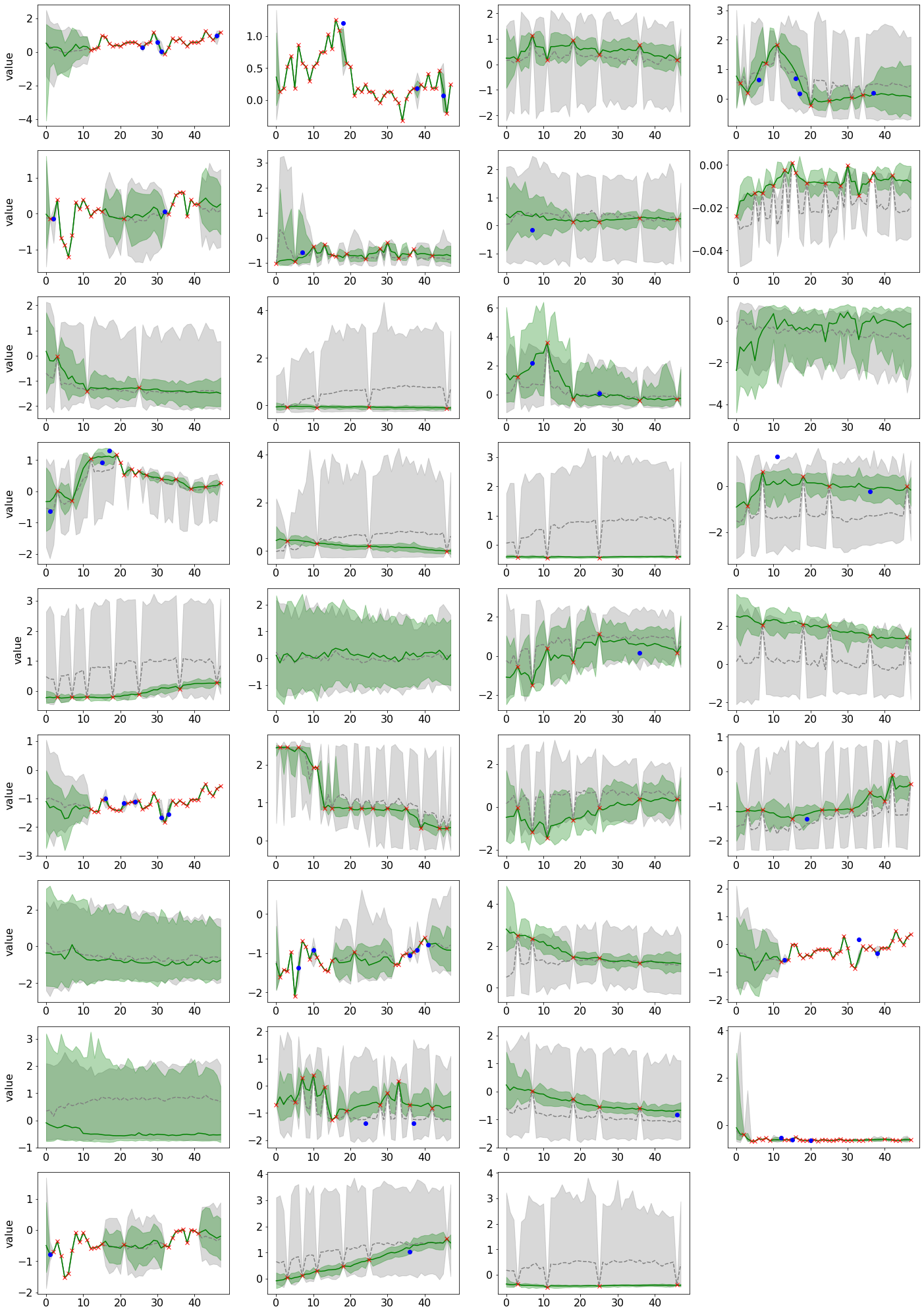}
        \end{center}
    \caption{Comparison of imputation between the unconditional diffusion model and \pname{} for the healthcare dataset (10\% missing). The result is for a time series sample with all 35 features. The red crosses show observed values and the blue circles show ground-truth imputation targets. Green and gray colors correspond to \pname{} and the unconditional model, respectively. For each method, median values of imputations are shown as the line and 5\% and 95\% quantiles are shown as the shade.} %
    \label{fig:example_10percent_uncond}
\end{figure}

\begin{figure}[htbp]
\centering

        \begin{center}
          \includegraphics[width=1.0\textwidth]{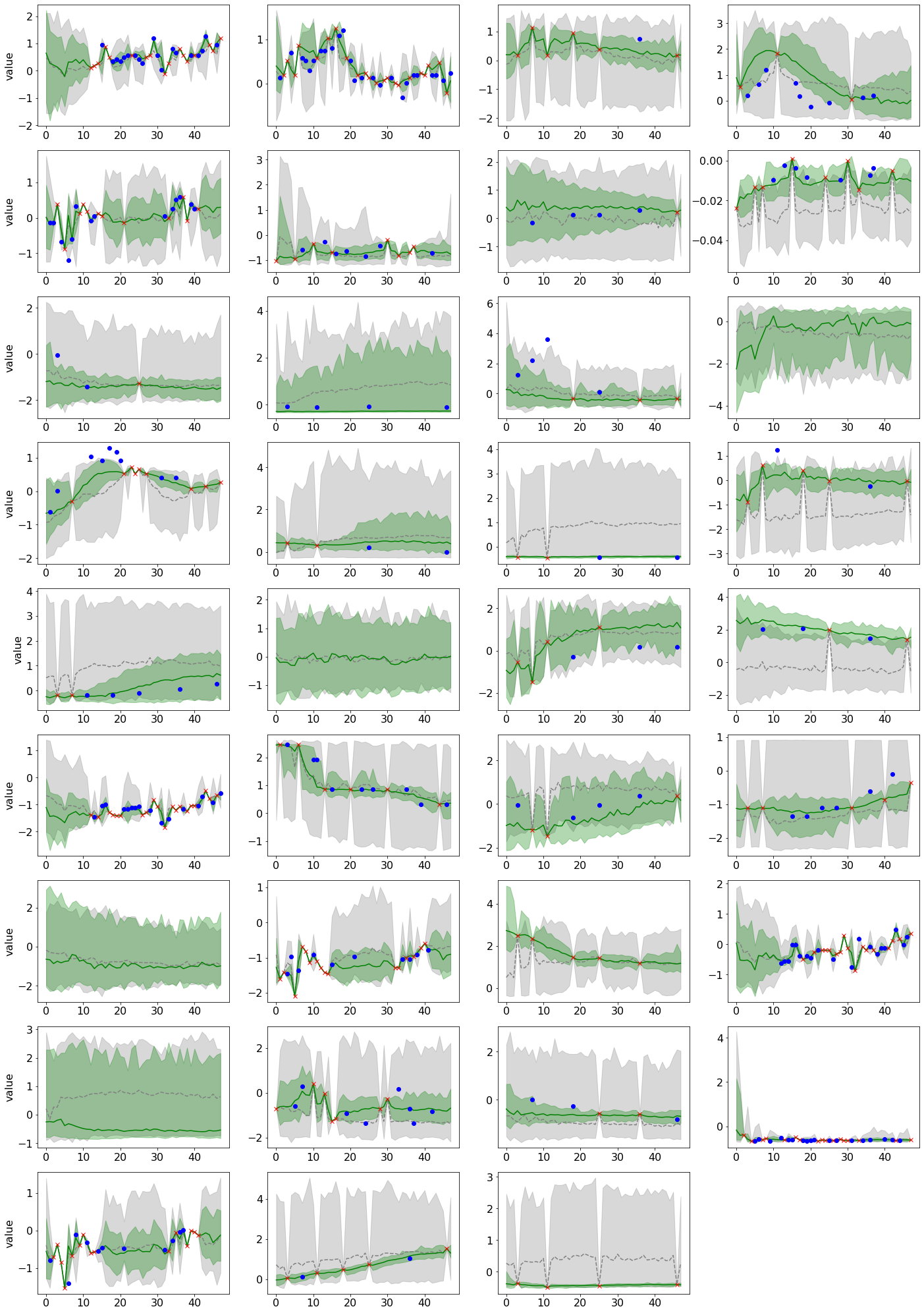}
        \end{center}
    \caption{Comparison of imputation between the unconditional diffusion model and \pname{} for the healthcare dataset (50\% missing). The result is for a time series sample with all 35 features. The red crosses show observed values and the blue circles show ground-truth imputation targets. Green and gray colors correspond to \pname and the unconditional model, respectively. For each method, median values of imputations are shown as the line and 5\% and 95\% quantiles are shown as the shade.} %
    \label{fig:example_50percent_uncond}
\end{figure}

\begin{figure}[htbp]
\centering

        \begin{center}
          \includegraphics[width=1.0\textwidth]{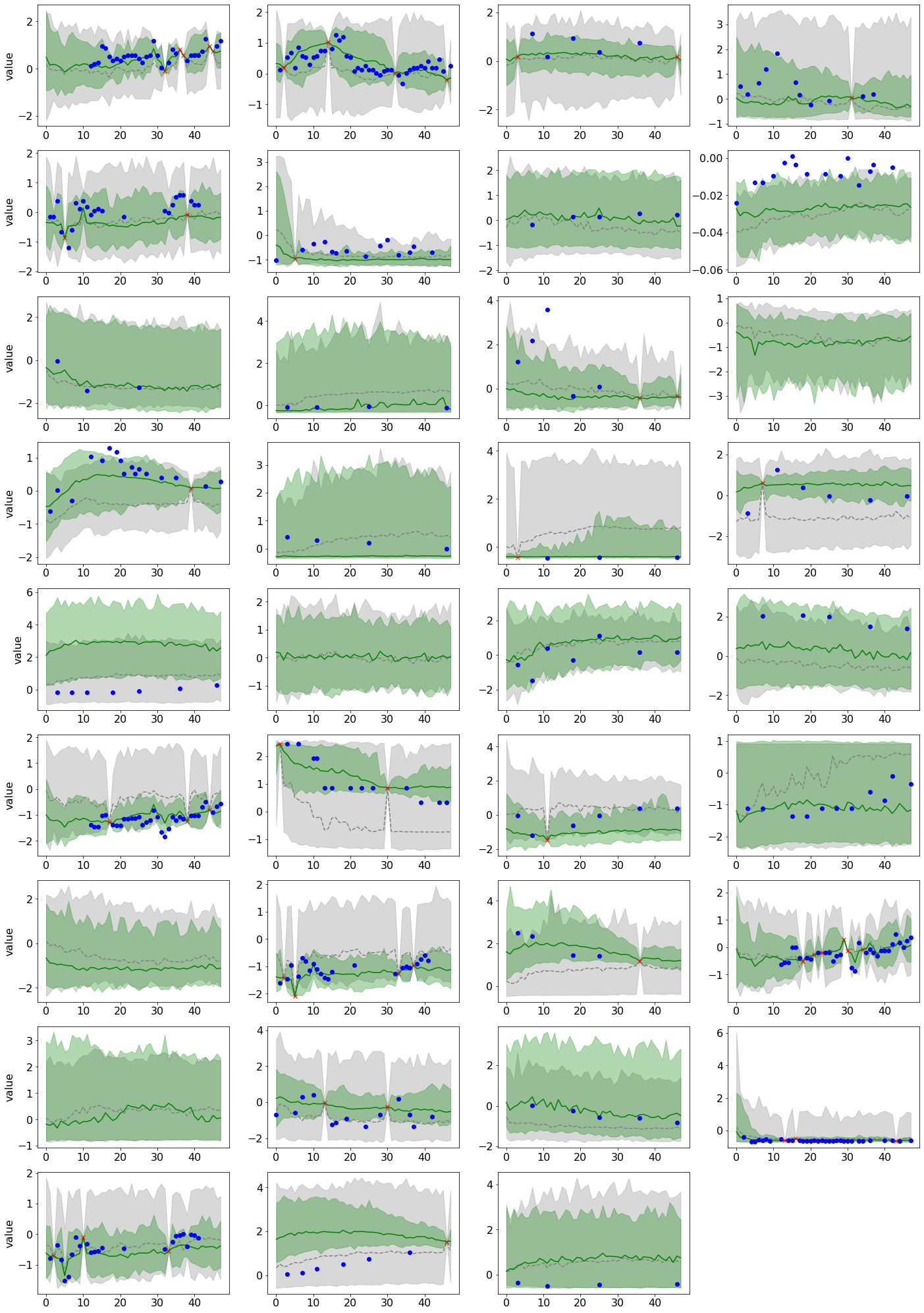}
        \end{center}
    \caption{Comparison of imputation between the unconditional diffusion model and \pname{} for the healthcare dataset (90\% missing). The result is for a time series sample with all 35 features. The red crosses show observed values and the blue circles show ground-truth imputation targets. Green and gray colors correspond to \pname{} and the unconditional model, respectively. For each method, median values of imputations are shown as the line and 5\% and 95\% quantiles are shown as the shade.} %
    \label{fig:example_90percent_uncond}
\end{figure}

\begin{figure}[htbp]
\centering
        \begin{center}
          \includegraphics[width=1.0\textwidth]{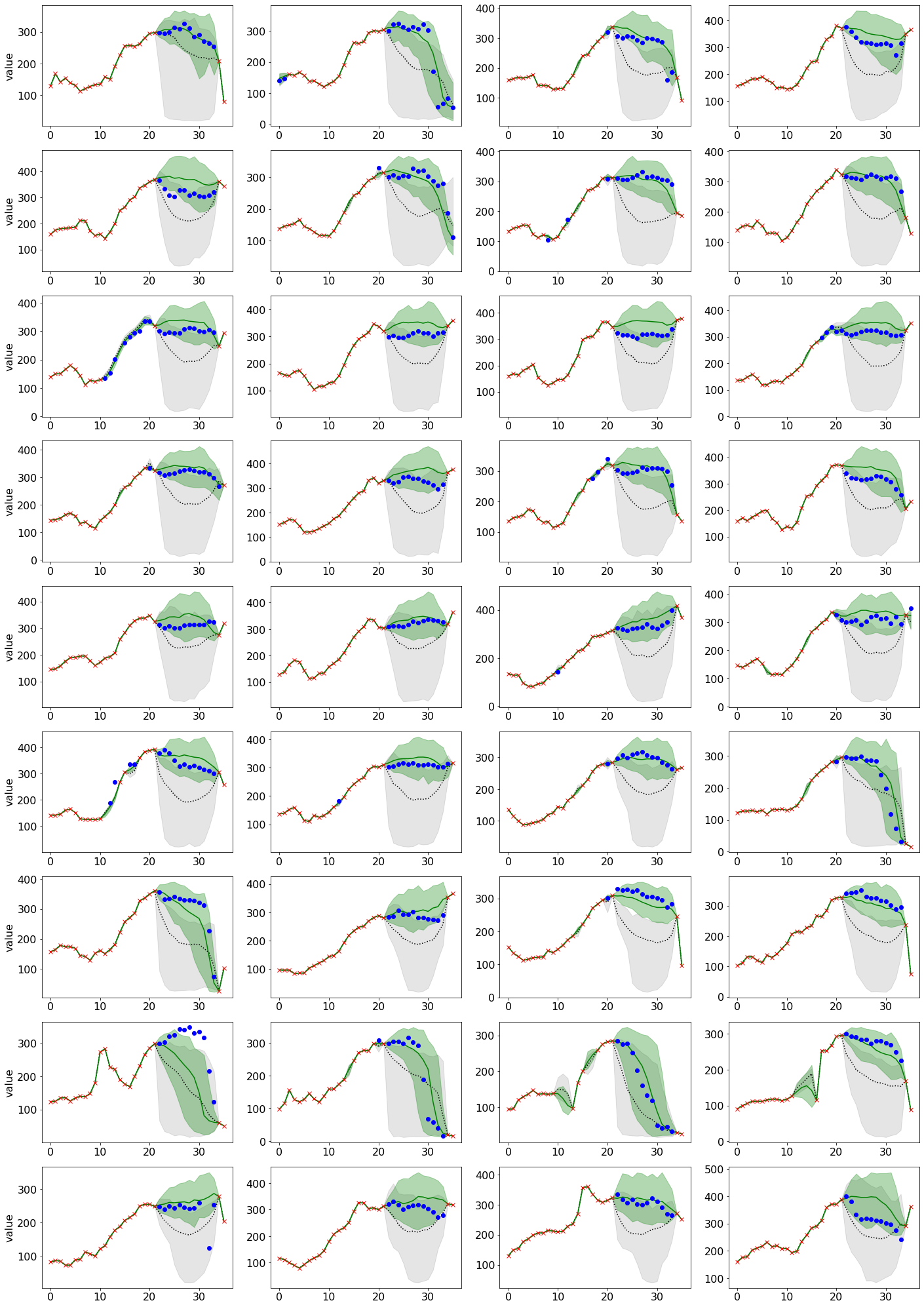}
        \end{center}
    \caption{Comparison of imputation between the unconditional diffusion model and \pname{} for the air quality dataset. The result is for a time series sample with all 36 features. 
    The red crosses show observed values and the blue circles show ground-truth imputation targets. Green and gray colors correspond to \pname{} and the unconditional model, respectively. For each method, median values of imputations are shown as the line and 5\% and 95\% quantiles are shown as the shade. } %
    \label{fig:example_airquality_uncond}
\end{figure}

In this section, we illustrate various imputation examples to show the characteristic of imputed samples. We pick a multivariate time series from the results of each experiment in \Secref{sec:experiments:imp} and show imputation results for all features of each time series in \Figref{fig:example_10percent} to~\ref{fig:example_airquality}. We compare \pname{} with GP-VAE in \Figref{fig:example_10percent} to~\ref{fig:example_airquality}. Note that the scales of the $y$ axis depend on the features. For the healthcare dataset with 90\% missing ratio in \Figref{fig:example_90percent}, while GP-VAE fails to learn the distribution, \pname{} gives reasonable probabilistic imputation for most of the features. For the air quality dataset in \Figref{fig:example_airquality}, \pname{} learns the dependency between features and provides more accurate imputation than GP-VAE. 
In \Figref{fig:example_10percent_uncond} to~\ref{fig:example_airquality_uncond}, we compare \pname{} with the unconditional diffusion model. 
In all figures, \pname{} tends to provide tighter uncertainty than the unconditional diffusion model. We hypothesize that it is due to the approximation discussed in \Secref{sec:background:imp}. Since the unconditional model approximates the conditional distribution by using noisy observed values, the estimated imputation become less confident than that with the conditional model.

\section{Potential negative societal impacts}
\label{append:negativeimpacts}
Since score-based diffusion models are generative models, our proposed model has negative impacts as well as other generative models. For example, the model can potentially memorize private information and be used to generate fake data.

\end{document}